\documentclass[preprint,10pt]{elsarticle}

\usepackage{amssymb}
\usepackage{amsmath}

\usepackage{amsfonts}
\usepackage{algorithmic}
\usepackage{algorithm}
\usepackage{array}
\usepackage[caption=false,font=normalsize,labelfont=sf,textfont=sf]{subfig}
\usepackage{textcomp}
\usepackage{stfloats}
\usepackage{verbatim}
\usepackage{graphicx}
\usepackage{cite}
\usepackage{wrapfig}
\usepackage[normalem]{ulem}
\usepackage{lineno}
\usepackage[colorlinks=false, pdfborder={0 0 0}]{hyperref}

\usepackage{lipsum}
\usepackage{mathtools}
\DeclarePairedDelimiter\abs{\lvert}{\rvert}

\usepackage[acronym, shortcuts, style=long]{glossaries}
\makeglossaries

\newacronym{ml}{ML}{Machine Learning}
\newacronym{ai}{AI}{Artificial Intelligence}
\newacronym{dl}{DL}{Deep Learning}
\newacronym{xai}{XAI}{eXplainable AI}

\newacronym{eo}{EO}{Earth Observation}
\newacronym{rs}{RS}{Remote Sensing}
\newacronym{scl}{SCL}{Scene Classification Layer}
\newacronym{ads}{ADS}{additional data sources}
\newacronym{rgb}{RGB}{Red-Green-Blue}
\newacronym{dem}{DEM}{digital elevation maps}
\newacronym{srtm}{SRTM}{Shuttle Radar Topography Mission}
\newacronym{aster}{ASTER}{Advanced Spaceborne Thermal Emission and Reflection Radiometer}
\newacronym{alos}{ALOS}{Advanced Land Observing Satellite}
\newacronym{ecmwf}{ECMWF}{European Center for Medium-Range Weather Forecasts}
\newacronym{sar}{SAR}{synthetic aperture radar}
\newacronym{lidar}{LiDAR}{Light Detection and Ranging}
\newacronym{fapar}{FAPAR}{Fraction of Absorbed Photosynthetically Active Radiation}
\newacronym{twi}{twi}{Topographic Wetness Index}
\newacronym{s2}{S2}{Sentinel-2}
\newacronym{l2a}{L2A}{Level-2A}
\newacronym{sits}{SITS}{satellite image time series}
\newacronym{gsd}{GSD}{ground sampling distance}
\newacronym{usda}{USDA}{United States Department of Agriculture}
\newacronym{fao}{FAO}{Food and Agriculture Organization}

\newacronym{ndvi}{NDVI}{normalized difference vegetation index}
\newacronym{evi}{EVI}{enhanced vegetation index}

\newacronym{rf}{RF}{random forest}
\newacronym{gpr}{GPR}{gaussian process regression}
\newacronym{svm}{SVM}{support vector machine}
\newacronym{mlp}{MLP}{multilayer perceptron}
\newacronym{dnn}{DNN}{deep neural network}
\newacronym{rnn}{RNN}{Recurrent neural network}
\newacronym{1d-cnn}{1D-CNN}{1-Dimensional convolutional neural network}
\newacronym{dfnn}{DFNN}{deep forward neural network}
\newacronym{lstm}{LSTM}{long short-term memory}
\newacronym{alstm}{ALSTM}{Attention-LSTM}
\newacronym{gbdt}{GBDT}{gradient-boosted decision tree}
\newacronym{pca}{PCA}{Principal Component Analysis}
\newacronym{mha}{MHA}{multi-head self-attention}
\newacronym{sdpa}{SDPA}{scaled dot-product attention}

\newacronym{mae}{MAE}{mean absolute error}
\newacronym{rmse}{RMSE}{root mean square error}
\newacronym{rrmse}{RRMSE}{relative root mean square error}
\newacronym{r2}{$\text{R}^2$}{coefficient of determination}
\newacronym{pp}{p.p}{percentage points}

\newacronym{cam}{CAM}{Class Activation Mapping}
\newacronym{gradcam}{Grad-CAM}{Gradient-weighted CAM}
\newacronym{shap}{SHAP}{SHapley Additive exPlanations}
\newacronym{lime}{LIME}{Local Interpretable Model-agnostic Explanation}
\newacronym{ig}{IG}{Integrated Gradients}
\newacronym{i*g}{I*G}{Input $\times$ Gradients}
\newacronym{svs}{SVS}{Shapley Value Sampling}
\newacronym{svcca}{SVCCA}{Singular Vector Canonical Correlation Analysis}
\newacronym{cca}{CCA}{canonical correlation analysis}
\newacronym{svd}{SVD}{singular value decomposition}
\newacronym{ar}{AR}{Attention Rollout}  
\newacronym{ga}{GA}{Generic Attention}  
\newacronym{wma}{WMA}{Weighted Modality Activation}

\newacronym{lofo}{LOFO}{leave-one-farm-out}
\newacronym{loyo}{LOYO}{leave-one-year-out}

\setacronymstyle{long-short}

\usepackage[usenames,dvipsnames]{color}
\usepackage[dvipsnames,table,xcdraw]{xcolor}
\usepackage{multirow}
\usepackage{changepage}
\usepackage{soul}
\usepackage{hyperref}
\usepackage{caption}
\usepackage{subcaption}
\usepackage{adjustbox}
\usepackage{enumerate}
\usepackage[shortlabels]{enumitem}

\begin{document}

\begin{frontmatter}

\title{Intrinsic Explainability of Multimodal Learning for Crop Yield Prediction} 
\author[rptu,dfki]{Hiba Najjar\corref{cor1}} 
\author[dfki]{Deepak Pathak} 
\author[dfki,bds]{Marlon Nuske} 
\author[rptu,dfki]{Andreas Dengel} 
\cortext[cor1]{Corresponding author. Email: najjar@rptu.de.}
\affiliation[rptu]{
            organization={RPTU Kaiserslautern-Landau},
            city={Kaiserslautern},
            country={Germany}}
\affiliation[dfki]{
            organization={German Research Center for Artificial Intelligence (DFKI)},
            city={Kaiserslautern},
            country={Germany}}
\affiliation[bds]{
            organization={Bundesanstalt für Landwirtschaft und Ernährung},
            city={Bonn},
            country={Germany}}

\begin{abstract}
    Multimodal learning enables various machine learning tasks to benefit from diverse data sources, effectively mimicking the interplay of different factors in real-world applications, particularly in agriculture. While the heterogeneous nature of involved data modalities may necessitate the design of complex architectures, the model interpretability is often overlooked. 
    In this study, we leverage the intrinsic explainability of Transformer-based models to explain multimodal learning networks, focusing on the task of crop yield prediction at the subfield level. 
    The large datasets used cover various crops, regions, and years, and include four different input modalities: multispectral satellite and weather time series, terrain elevation maps and soil properties. 
    Based on the self-attention mechanism, we estimate feature attributions using two methods, namely the \gls{ar} and \gls{ga}, and evaluate their performance against Shapley-based model-agnostic estimations, \gls{svs}. 
    Additionally,  we propose the \gls{wma} method to assess modality attributions and compare it with \gls{svs} attributions.  
    Our findings indicate that Transformer-based models outperform other architectures, specifically convolutional and recurrent networks, achieving $\text{R}^2$ scores that are higher by 0.10 and 0.04 at the subfield and field levels, respectively.
    \gls{ar} is shown to provide more robust and reliable temporal attributions, as confirmed through qualitative and quantitative evaluation, compared to \gls{ga} and \gls{svs} values.
    Information about crop phenology stages was leveraged to interpret the explanation results in the light of established agronomic knowledge.
    Furthermore, modality attributions revealed varying patterns across the two methods compared. For instance, \gls{svs} estimated the contribution of satellite data at 89.5\% on average, whereas the \gls{wma} method provided a significantly lower estimate of 29.4\%. These results call for further analysis and quantitative evaluations.
    Overall, this work contributes to the growing body of research aiming at enhancing the interpretability of multimodal networks in challenging data-rich contexts in agriculture and remote sensing. 

    The implementation details of the model and attribution methods will be open-sourced upon paper acceptance.

\end{abstract}

\end{frontmatter}

\section{Introduction} \label{sec:intro}

    Yield prediction is a pivotal task for promoting sustainable agriculture and advancing digital farming. 
    Accurate yield estimates play a key role in completing historical yield records and addressing potential data gaps. For example, such yield ``\textit{simulation}", or ``\textit{imputations}", are particularly valuable for government agencies such as the \gls{usda} and Eurostat, as well as organizations like the \gls{fao}, to publish agricultural yield data at regional and national levels. Additionally, universities with agricultural research programs rely on historical yield records to assess the impact of weather and other factors on crop yields. Satellite data providers also benefit from yield estimates, as they can incorporate them into agricultural data offerings alongside primary satellite imagery.
    On the other hand, forecasts of crop yield can support local efforts to enhance agricultural profitability and inform regional strategies. This can be achieved by comparing estimated yields with current and future demands, subsequently adjusting import and export plans to ensure food security.
    In this manuscript, we focus on the first use case of yield simulation (filling gaps in historical records) but use the term ``yield prediction" throughout the paper to align with common usage in the literature.

    As highlighted by \citet{weiss2020remote}, crop yield prediction is a more complex and challenging task compared to predicting other agronomic traits, such as canopy height and green area index. 
    To capture some of the complex and multidimensional factors influencing crop yields, \gls{ml} and multi-modal learning have been increasingly used to integrate a variety of data sources, such as satellite imagery, weather data, and soil information \citep{mena2024common}. In fact, it was shown that models fusing data from different modalities outperform their uni-modal counterparts both intuitively and provably \citep{huang2021makes}.

    Due to the importance of model accuracy, most work in the multimodal learning literature primarily focuses on designing complex architectures and optimizing performance, with limited emphasis on the interpretability of these models \citep{rahate2022multimodal,joshi2021review}. Given the often opaque nature of multimodal architectures, understanding how different modalities contribute to model predictions is crucial. 
    In this context, intrinsic interpretability methods, which provide explanations directly tied to the model’s internal components, offer a promising alternative to traditional model-agnostic explanation approaches that treat the model as a block box \citep{rudin2019stop}. 
    Intrinsic explanations are inherently more faithful and less prone to errors introduced by surrogate models \citep{lime_ribeiro2016should, shap_lundberg2017unified, molnar2020interpretable}. 
    The need for intrinsic interpretability is especially relevant in \gls{rs} applications, where multiple data modalities, such as satellite imagery, climate and weather data, and topographical maps, are commonly used to predict complex environmental and agricultural phenomena \citep{mena2024common, li2022deep, gunther2024explainable, russwurm2020self}. 
    This is particularly important in agriculture, where the model's decisions must be trusted, as they have long-term implications for related fields such as ensuring food security and informing national exchange policies \citep{weiss2020remote}.
    In other words, after explaining a model and validating its reasoning process, it becomes more reliable when deployed for the intended end-task.
    Accordingly, our work explores the transparency and intrinsic interpretability of multimodal networks for crop yield prediction. Specifically, this study investigates the following research questions: 
    
        \begin{enumerate}[label=\bfseries{RQ\arabic*:}, ref=\bfseries{RQ\arabic*},leftmargin = 4em]
            \item \label{rq1} Which multimodal network architecture performs best for the task of yield prediction, given the four modalities used in this study?
            \item \label{rq2} What can the analysis of the intermediate representations reveal?
            \item \label{rq3} Which method for estimating temporal attributions is most reliable?
            \item \label{rq4} Can the temporal attributions provide agronomically relevant insights?
            \item \label{rq5} Which modality attribution method is most reliable?\\
        \end{enumerate}
        
\begin{figure}[t]
        \centering
        \adjustbox{center=\textwidth}{\includegraphics[width=1.2\linewidth]{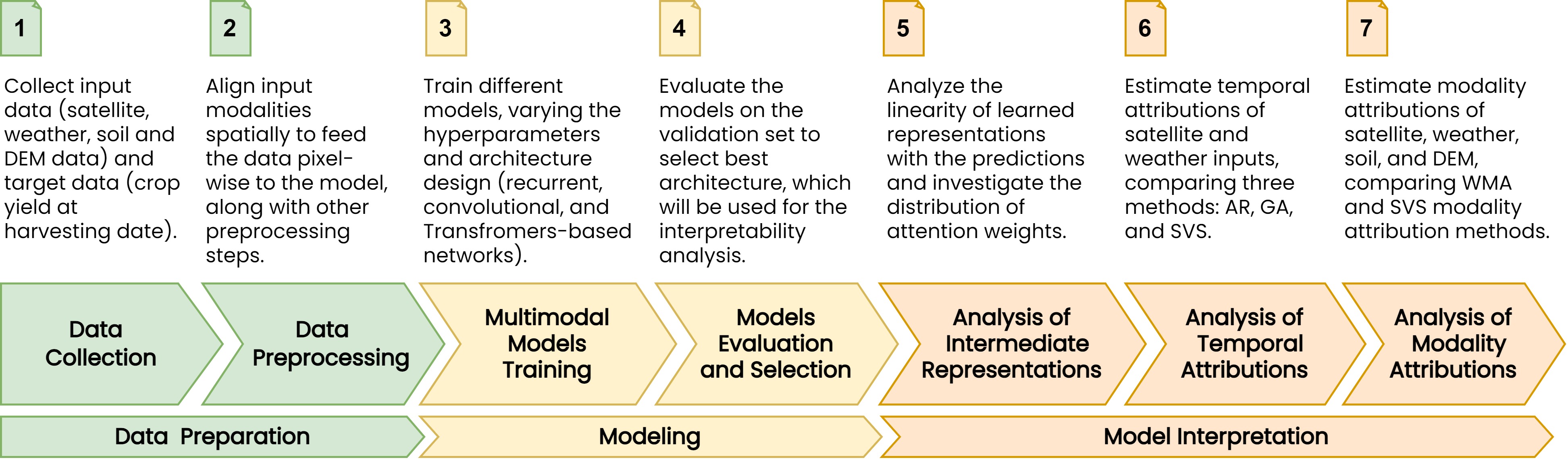}}
        \caption{General workflow and key steps of the study.
        }
        \label{fig:workflow}
    \end{figure}

    Figure \ref{fig:workflow} illustrates the overall workflow and key steps of the study.
    The remainder of this paper is organized as follows: Section \ref{sec:rel_work} provides an overview of prior work on explainability in multimodal learning networks, with a particular focus on leveraging self-attention mechanisms for explainability in agriculture and \gls{rs} applications. 
    Sections \ref{sec:modeling} and \ref{sec:xai} outline our modeling and explainability methodologies, while the crop yield datasets are described in Section \ref{sec:data}.
    All results are presented in Section \ref{sec:results}, including 
        model evaluation at subsection \ref{ssec:model_eval},
        analysis of intermediate representations at subsection \ref{ssec:prob_repr},
        investigation of temporal attributions at subsection \ref{ssec:temp_attr} and their correlation with specific weather events in subsection \ref{ssec:weather_events},
        and the introduction of a modality importance estimation technique in \ref{ssec:mod_imp}.
    Finally, we gather the main insights of our extensive study in the Discussion, i.e. Section \ref{sec:discussion}, and summarize our findings in the Conclusion, i.e. Section \ref{sec:conclusion}.

\section{Related Work}\label{sec:rel_work}

        Explainability in multimodal learning networks has gained increasing attention as these models allow the combination of diverse data types, yet the difficulty of this task results in complex architectures and threatens the interpretability of their decision-making process \citep{joshi2021review}. 
        Feature attribution techniques, such as \gls{shap} \citep{shap_lundberg2017unified} and Integrated Gradients \citep{ig_sundararajan2017axiomatic}, are model agnostic \gls{xai} techniques, and can thus easily be applied to multimodal networks. 
        Recently, graph-based explainability methods have been proposed to model inter-modality dependencies more comprehensively \citep{ghosh2019generating,gaur2021semantics}. 
        Other methods leverage attention mechanisms to highlight the importance of different modalities and their interactions, yet such applications often only visualize the attention weights of certain input samples, which provides very limited insights into the more general understanding of the model \citep{ghosal2018contextual,tsai2019multimodal}. 
    
        The \gls{rs} field is particularly rich in modalities, making the explainability of multimodal learning in this context crucial for sensitive applications including agriculture \citep{gunther2024explainable}. 
        One particularly challenging \gls{rs} agricultural application is crop yield prediction, due to the involvement of numerous interdependent factors.
        Moreover, ground truth yield data is typically provided by government agencies, national organizations, and international institutions. However, such data is often available only at coarse spatial resolutions, aggregated at county or state levels. While some studies have access to field-level yield aggregates, leaving the sub-field level relatively underexplored  \citep{leukel2023machine, muruganantham2022systematic, nevavuori2019crop}.
        Beyond yield data granularity, the application of multimodal learning can be further classified into studies that employ either an early-fusion approach \citep{cai2019integrating, gavahi2021deepyield, wang2020winter, cao2021integrating} or those that apply a modality-specific encoding of the data before applying an intermediate or late fusion of the learned representations \citep{pathakPredictingCrop2023, ma2023field, yang2019deep, maimaitijiang2020soybean, jeong2022predicting, mena2025adaptive}.
        
        Several studies have explored the explainability of the yield prediction task using multiple input modalities, including \gls{rs} data, primarly focusing on the case of early data fusion. These studies often employ model-agnostic explanation methods \citep{martinez-ferrerCropYield2021,bromsCombinedAnalysis2023,huberExtremeGradient2022,paudel2023interpretability,wang2023learning,isik2023interpretable,najjar2025explainability}, as described in \citet{najjar2025explainability}.
        Few other cases used intrinsic interpretability \citep{farmonovCombiningPlanetScope2023,celik2023explainable,wolaninEstimatingUnderstanding2020}.
        For instance, 
            \citet{farmonovCombiningPlanetScope2023} train a \gls{rf} to predict wheat yield at the pixel level and estimate feature importances using the node impurity method inherent to ensemble tree-based models. 
            \citet{celik2023explainable} train generalized additive models to predict cotton yield, and the learned model parameters to evaluate the interactions between the features, their importance, and their interpretation.
            \citet{wolaninEstimatingUnderstanding2020} train a convolutional network to predict wheat yields in the Indian Wheat Belt and apply regression activation maps to visually explain the model and identify significant time steps.
        In our work, we focus on the intermediate fusion approach in multimodal networks and the comparative analysis of different explanation methods.

        Taking a closer look at the use of self-attention mechanisms to leverage their inherent interpretability in \gls{rs}, researchers have explored this approach for several tasks, including crop classification \citep{khan2024transformer, xu2021towards, russwurm2020self, garnot2020satellite, obadic2022exploring}, land cover classification \citep{kim2022federated, meger2022explaining}, water quality monitoring \citep{pyo2021cyanobacteria}, and target detection \citep{zhou2019local}. However, the analysis of self-attention mechanisms for \gls{xai} in these studies is often limited, with little focus on in-depth interpretability. For instance, 
            \citet{khan2024transformer} compared two Transformer-based architectures for land cover classification, employing multiple post-hoc explanation techniques to elucidate the predictions. However, the self-attention mechanism was not leveraged for intrinsic model explanation.
            \citet{kim2022federated} extracted multiple attention maps from a convolutional network embedded in a satellite's on-board system, automatically identifying samples with inconsistent maps. These samples were then sent to a ground station for correction by expert annotators before being returned to the satellite to update the model. While this work provides a framework that utilizes attention maps as a tool for improving the model in a weakly supervised manner, it is restricted to computer vision tasks and focuses primarily on local explanations (i.e., explaining individual predictions without generalizing insights across multiple samples).
            \citet{xu2021towards} trained a \gls{lstm} network combined with an attention layer (thus creating an \gls{alstm} model) and a Transformer model for crop classification, and subsequently analyzed raw attention weights to provide explanations. While their analysis was primarily descriptive, further processing of the attention weights could have revealed deeper insights into the model, as we will demonstrate in our work.

        In the context of the yield prediction task, multiple studies have explored the impact of the time- and region-wise drifts on the model performance \citep{helberCropYield2023,helber2024operational}, while others used attention-based models to enhance task accuracy \citep{mena2025adaptive, inderka2024convolutional, krishnan2024sugarcane, qiao2023kstage, lin2023mmst, junankar2023wheat}.
        Nevertheless, we could identify only a single study which has explicitly focused on explaining such attention-based models: \citet{tian2021deep} trained an \gls{alstm} model to predict winter wheat yield at the county level in central China. Their approach involved an early fusion of two vegetation indices and two climate-related features. The yield data covered only 22 counties from Shaanxi province, with spatial and temporal resolutions of 500m and four growth stages, respectively.
        However, this study does not leverage the attention mechanism for inherent explainability and instead relies on post-hoc methods.
        
        Our work demonstrates how the attention mechanism, particularly in Transformer-based models, can be leveraged to enhance the intrinsic interpretability of multimodal networks. 
        This is different from prior studies, which either rely on post-hoc explanation methods or use less advanced modality fusion techniques. 
        We conduct our analysis on the yield prediction task, contributing in the following four aspects:

        \begin{enumerate}
              
            \item \textbf{Multimodal learning}: To accurately predict crop yield, we incorporate four modalities (i.e. time series of satellite and weather data, terrain elevations, and soil properties), processed individually before applying an  intermediate fusion of learned representations.
            \item \textbf{Sub-field yield modeling}: We use high-resolution yield records as target values and compare the results of main analysis across three countries and four crop types.
            \item \textbf{Model interpretability}: We leverage the inherent interpretability of the attention mechanism to explain Transformer-based multimodal networks through extensive analysis.
            \item \textbf{Post-hoc vs. intrinsic}: We compare the performance of intrinsic explanation methods against post-hoc model-agnostic techniques.
         
        \end{enumerate}

\section{Modeling}\label{sec:modeling}
    To address \ref{rq1}, this section outlines the models used for crop yield prediction based on pixel-wise processing of spatially aligned modalities. We test various neural network architectures for encoding individual modality information and fusing the learned representations.

    \subsection{Modality Encoders}
\begin{figure*}[t]
            \centering  
            \includegraphics[width=\linewidth]{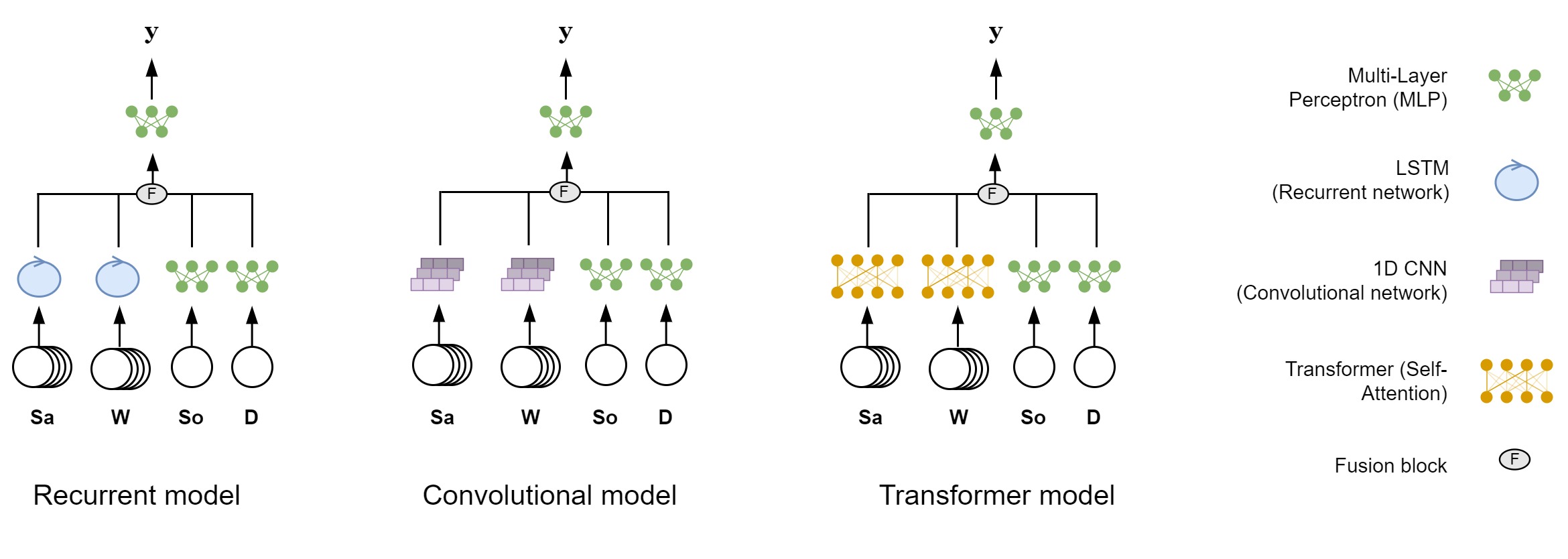}
            \caption{Multimodal architectures with concatenation fusion for yield prediction.
            The abbreviations used for the input modalities are as follows: Sa [Satellite], W [Weather], So [Soil], and D [\gls{dem}]. These modalities are described in Section \ref{sec:data}.
            }
            \label{fig:model_arch}
        \end{figure*}

        Depending on the modality's nature, i.e., static or temporal, we use different neural network architectures to encode its representation. For static modalities, such as the terrain elevations (i.e. \gls{dem}) and soil properties, we use \glspl{mlp}. For temporal modalities, such as satellite and weather data, we test three different types of architectures: 
        recurrent networks,  convolutional networks, and Transformers. 
        Each of these modality encoders is expected to produce a representation denoted as $h\in \mathbb{R}^d$.
        Figure \ref{fig:model_arch} depicts the different architecture types used. 
        
        \subsubsection{Multi-Layer Perceptron}
            \glspl{mlp} are a type of artificial neural network where information flows from the input layer to the output layer, without any loops or cycles. \glspl{mlp} extract features by learning high-level representations through layers of neurons, each performing a weighted sum followed by a non-linear activation function. In our implementation, we use two fully connected layers: the first (intermediate) layer has a dimension of $2d$, and the second (output) layer has a dimension of $d$, which returns the modality representation. Batch normalization and the ReLU activation function are applied after the first layer.
        
        \subsubsection{Recurrent Networks}
            \glspl{rnn} are inherently capable of handling temporal data. They process one temporal instance at a time, learning to predict outputs and maintain a hidden state at each step. The hidden state is optimized to focus on important information while discarding irrelevant or redundant data. In our implementation of the \gls{rnn}, we use a stack of two \gls{lstm} cells \citep{hochreiter1997long} with a dropout rate of 0.3, followed by a linear layer which transforms the \gls{lstm} output at the final time step to a dimension of $d$. Before applying the linear layer, we include batch normalization to improve training stability.

            We also explore another \gls{rnn} variant based on ALSTM \citep{tian2021deep}, which aggregates outputs from all time steps using a weighted combination, rather than relying solely on the final time step. The weights are computed using a form of scaled dot-product attention \citep{vaswani2017attention}.
        
        \subsubsection{Convolutional Networks}
            \glspl{1d-cnn} are commonly used for processing sequential data, such as time series or natural language, by applying convolutional filters across a one-dimensional input.
            Unlike \glspl{rnn}, which process one time step at a time, \glspl{1d-cnn} use convolutional filters to capture patterns or features in a hierarchical manner along the temporal dimension of the input. Our implementation follows the feature extraction approach in \citep{pelletier2019tempCNN}, with the modification of using a linear layer at the end instead of a SoftMax layer to produce a modality representation of dimension $d$.
        
        \subsubsection{Transformers}\label{sec-transformers}
\begin{figure}
                \centering
                \includegraphics[width=\linewidth]{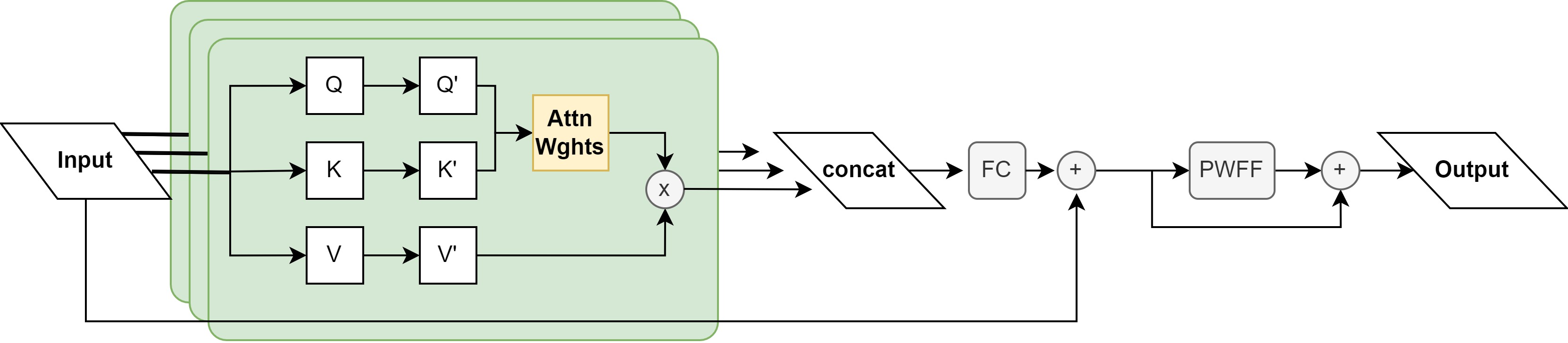}
                \caption{Schematic representation of a single Transformer layer composed of three heads. Q, K, and V stand for the query, key and values used at the attention mechanism, which are all a duplicate of the input sequence. FC refers to 'fully connected layer', while PWFF stands for the 'pixel-wise feed-forward' network.}
                \label{fig:trsf_layer}
            \end{figure}

            Transformers are highly effective for modeling temporal data due to their ability to use self-attention mechanisms \citep{bahdanau2016end, vaswani2017attention} to capture long-range dependencies within the input sequence. Unlike \glspl{rnn} and \glspl{1d-cnn}, which process data sequentially or locally, Transformers attend to all time steps simultaneously, allowing them to more effectively capture complex temporal patterns. 
            The input features are first passed through a linear embedding layer, which transforms each time step into a token of size $d$, while a learnable regression token, similar to \textit{class token} in \citep{devlin2018bert, dosovitskiy2020vit}, is added to interact with all time steps. 
            Before adding the regression token and feeding the data to the Transformer layers, positional encoding is applied based on the date of the time step, as suggested in \citep{vaswani2017attention}. Specifically, we use two calendar years, covering the crop season, and for each time step, we calculate the number of days elapsed from the beginning of the first year to determine its index. This positional encoding follows the approach of \citep{vaswani2017attention}, except we use the index calculated in days as described.
            The transformed input is then processed through multiple layers of Transformer encoders, each consisting of \gls{mha} and position-wise feed-forward networks. In each Transformer layer, the input undergoes layer normalization before being processed through the \gls{mha}. The output from the \gls{mha} layer is added back to the input via a residual connection, followed by a second normalization step. A position-wise feed-forward network is then applied, with its output also added through residual connections. This process is repeated across several Transformer layers, with the final modality representation derived from the output of the regression token. Figure \ref{fig:trsf_layer} illustrates the architecture of a single Transformer layer with three heads.

        \subsubsection{Fusion and Regression} 
            Given the heterogeneous nature of the input modalities usually interacting in agricultural applications and \gls{rs}, intermediate-level fusion is well-suited for our study, as opposed to input- or output-level fusion \citep{liang_mml_fusion_2024, mena2024common}. 
            Each modality is first processed by a dedicated encoder, which maps the input into a feature representation of dimensionality $d$. The learned representations from all $m$ modalities are then fused via simple concatenation along the feature dimension, and the resulting fused representation has a dimensionality of $d' = m \times d$. Finally, this fused representation is passed through a linear regression layer, which maps the $d'$-dimensional feature vector to a scalar output for yield prediction.

    \subsection{Model Finetuning}\label{ssec:model_ft}
        The different model architectures incorporate multiple hyperparameters that can influence model performance. 
        We experimented with various configurations of hidden sizes, number of layers and number of attention heads to optimize performance for the yield prediction task, as we describe in \ref{app:model_ft1}. 
        For this purpose, the dataset was split into training, validation, and test sets, with the validation set used to select the best network configuration, and the test set used to evaluate and report the model's performance on unseen data.

    \subsection{Model Training}
        The models were trained using mini-batch stochastic gradient descent with the Adam optimizer and decoupled weight decay (AdamW) \citep{loshchilov2017decoupled}. We employed a learning rate scheduler that begins with a linear warm-up for 5 epochs, followed by cosine decay for 50 epochs \citep{loshchilov2022sgdr}. Early stopping was implemented to stop training when the validation loss did not decrease for 10 consecutive epochs. Details of the hyperparameter finetuning method for used optimizer, learning rate scheduler, weight decay, and batch size are provided in \ref{app:model_ft1}.

\section{Explainability}\label{sec:xai}
    An central contribution of our study is the interpretation of multimodal networks, as outlined in \ref{rq2}-\ref{rq5}. 
    In the following, we describe the various tools used to understand learned representations and explain the yield prediction model, with particular emphasis on Transformer-based architectures.
        
    \subsection{Attention layers dynamic}\label{ssec:lp}
        To better understand the roles and dynamics of the intermediate layers, we use linear classifier probes \citep{alain2016understanding}. 
        In practice, linear probes consist of linear regressors that take as input the latent features learned by an intermediate layer of the trained model and learns to predict the corresponding yield value, as predicted by the model. High accuracy of this regressor suggests a linear separability of the features at the examined layer. By comparing the accuracy of linear probes across successive layers, we can verify how the learned features gradually become more separable across different modality encoders.
        
    \subsection{Self-attention mechanism}\label{ssec:attn}
        Since the introduction of attention mechanisms in the literature, many have seen the opportunity to use the weights for explaining neural networks \citep{vaswani2017attention, russwurm2020self, xu2021towards}. 
        Indeed, the attention weights link the input to the subsequent layers of the network, allowing the model to focus on relevant parts of the input, and this link is used to interpret the model reasoning behind individual predictions. 
        
        Temporal attentions are extracted from the attention weights to identify the time steps prioritized by the model as it makes its final prediction.
        Let $A^l \in \mathbb{R}^{T \times T}$ denote the attention matrix of layer $l$, where $T$ is the number of time steps and $A^l_{j,t}$ is the attention weight assigned to time step $t$ by time step $j$.
        To get temporal attentions $S^l = \{ S^l_1, S^l_2, \cdots, S^l_T\}$ , we first compute the total attention received by each time step $t$ at layer $l$ from all time steps. This value is then normalized by dividing by the total number of time steps $T$, resulting in a probability distribution:
        \begin{equation}\label{eq:temp_attn}
            S^l_t = \frac{1}{T} \sum_{j=1}^{T} A^l_{j,t}, \quad \forall t \in \{1, \dots, T\}.
        \end{equation}

        This is not to be confused with the summation over each row of the attention matrix, where the weights form a probability distribution over time steps due to the row-wise SoftMax normalization \citep{vaswani2017attention}, and thus: 
        $$
        \sum_{t=1}^{T} A^l_{j,t} = 1, \quad \forall j \in \{1, \dots, T\}.
        $$
        The time series $S^l$ of temporal attentions is computed for each layer and head of the Transformer encoders to understand how attention is distributed throughout the network.
        For the final layer $L$, only the attention weights associated with the regression token $r$ are evaluated, as all other time steps are excluded from subsequent processing by the model:
        $$
        S^L_t = A^L_{r,t}, \quad \forall t \in \{1, \dots, T\}.
        $$

        To evaluate the information content within temporal attentions, we use Shannon entropy as defined in the foundational work by \citet{shannon1948mathematical}. 
        The entropy is computed for each time series of temporal attention $S^l$ at every layer of each Transfromer encoder.
        Given that the temporal attention values are normalized in Equation \ref{eq:temp_attn}, we consider their range within [0,1] and divide this interval into 100 bins for computing the entropy. This ensures comparability across modalities. 
        Low entropies indicate that the model focuses its attention on specific time steps, while high entropies suggest it spreads attention more evenly.

    \subsection{Temporal Attribution}
        To assess and compare the influence of different time steps on the model predictions, we use three feature attribution methods, as we describe in the following.
        
        \subsubsection{\acrfull{ar}} 
            In a multi-head multi-layer Transformer block, each sample generates multiple attention weight matrices. Direct analysis of each matrix can be time-consuming and might not easily reveal the inner workings of the model.
            Additionally, as we progress through deeper layers of the model, the identifiability of individual time steps decreases, resulting in increasingly mixed information. Consequently, direct probing of attention weight matrices for explainability becomes impractical.
            Therefore, to trace the information propagated from the input layer to the final embeddings of each Transformer block, we employ \gls{ar} \citep{abnar2020quantifying}. This method treats attention weights as proportion factors and iteratively multiplies the attention weight matrices of the multiple attention layers. The resulting matrix encodes the attention distributions of the entire Transformer block and can thus serve as a reliable basis for explanation.
            In our analysis, we specifically focus on the attention weights corresponding to the regression token.
            
        \subsubsection{\acrfull{ga}} 
            Another approach that leverages the internal workings of the Transformer model and facilitates its interpretation is \gls{ga}  \citep{chefer2021generic}. 
            Unlike \gls{ar}, which only uses the attention matrices, \gls{ga} propagates information backward from the final output through the last Transformer layer and subsequently through all preceding layers using gradients.
            As with \gls{ar}, our analysis focuses on the resulting weights attending to the regression token.
           
        \subsubsection{\acrfull{svs}} 
            Shapley values \citep{shapley1953value}, a concept derived from cooperative game theory, are commonly applied in the field of \gls{xai} as a model-agnostic and post-hoc feature attribution method. 
            In contrast to \gls{ar} and \gls{ga}, Shapley values are estimated using only the model and input samples, treating the model as a black box. 
            This is achieved by masking certain features, passing the modified sample through the model, and measuring the change in prediction. 
            In our implementation, we mask entire time steps, replacing them with baseline values computed as the mean of each masked variable across the dataset.
            Exact Shapley values quantify the average marginal contribution of each feature by evaluating its impact across all possible subsets of features.
            To mitigate the high computational cost of computing exact Shapley values, we employ their approximation technique \gls{svs} \citep{svs_strumbelj2010efficient}. 
            \gls{svs} has demonstrated superior robustness in terms of sensitivity and fidelity compared to other attribution methods on similar yield prediction tasks \citep{yeh2019fidelity, najjarFeatureAttribution2023}.

        \subsubsection{Attribution Evaluation}               
            Two evaluation metrics, \textit{sensitivity} and \textit{infidelity}, are used to assess the robustness of feature attributions estimated by the \gls{ar}, \gls{ga}, and \gls{svs} methods \citep{yeh2019fidelity}. Each metric provides a single score per data sample to assess the quality of the computed attributions.
                The sensitivity score measures how much the explanation varies when the input $X$ is perturbed slightly, indicating the stability of the attribution under small input changes.  
                The infidelity score evaluates the expected difference between (i) the dot product of the input perturbation and the attributions vector of $\mathbf{X}$, and (ii) the resulting difference in model output between $\mathbf{X}$ and the perturbed input $\mathbf{X}^\prime$.  

    \subsection{Modality attribution}\label{ssec:xai_mod}
        Beyond feature attributions, we investigate two modality attribution methods in order to compare the relative impact of the different modalities on model predictions. The first method is derived from the Shapeley-based feature attributions, \gls{svs}, while the second one is a newly proposed method, based on the inner parameters of the model, as we describe in the following. 
        
        \subsubsection{\gls{svs}-based modality importance}
            \gls{svs} can estimate the contribution of each individual input feature to the model's predictions. 
            To get the relative importance of different data modalities, we aggregate the absolute \gls{svs} scores per modality. 
            Specifically, for each pixel, we compute the importance score of each input feature by taking the absolute values of the \gls{svs} scores, which are then summed separately for each modality. 
            To ensure comparability, we subsequently scale the modality scores such that they sum to one. 

        \subsubsection{\acrfull{wma}}
            Since the multimodal networks described in Section \ref{sec:modeling}  use a concatenation-based fusion block followed by a linear layer, we propose to exploit this structure to infer modality attributions. 
            We can rewrite the final prediction $\hat{y}_i$ of sample $i$ as the weighted combination of the modality activations 
            $\mathbf{z}_i = \operatorname{concat} \left( \mathbf{z}_i^m \right)$, with $m \in \{  \text{satellite ($sa$), weather ($w$), soil ($so$), dem} \} $ and infer modality relevance scores $ \mathcal{R}_i^{m}$:
            
            $$
                \hat{y}_i = \mathbf{w}.\mathbf{z}_i + b  = \sum_m \mathbf{w}^{m}.\mathbf{z}_i^{m} + b = \sum_m \hat{y}_i^m +b, 
            $$
            $$
                \mathcal{R}_i^{m} = \abs{\frac{\hat{y}_i^{m}}{\hat{y}_i-b}}, 
            $$

            where $\mathbf{w} = \operatorname{concat} \left( \mathbf{w}^{sa}, \mathbf{w}^w, \mathbf{w}^{so}, \mathbf{w}^{dem} \right)$ and $b$ are the weights vector and bias of the final regression layer, respectively.                
            This approach can be viewed as an alternative to \gls{cam} and \gls{gradcam} methods \citep{cam_zhou2016learning,gradcam_selvaraju2017grad}, which are widely used for explaining classification tasks in computer vision. However, while \gls{cam} and \gls{gradcam} are specifically designed for convolutional networks operating on a single modality, our method is applicable to any multimodal regression task using a concatenation fusion mechanism and a \gls{mlp} as a regression head. Furthermore, it can be extended to various differentiable fusion strategies and regression heads using gradient-based techniques, similar to \gls{gradcam}.

\section{Data}\label{sec:data}
    To predict the yield, we use four modalities, including satellite data, weather information, soil properties, and \gls{dem}.
    Yield measurements, originally collected by combine harvesters, are preprocessed and used as target values.
    In the following subsections, we provide detailed descriptions of each data component used in our study.

    \subsection{Yield data}\label{app:yield_data}
    
        \begin{table}[]
            \centering
            \caption{Yield data description. We train different models for each country-crop pair.}
            \small
            \label{tab:yield_datasets}
            \begin{tabular}{cccccc}
                \hline
                \textbf{Country} & \textbf{Crop} & \textbf{Years} & \textbf{\# Farms} & \textbf{\# Fields} & \textbf{\# Pixels} \\ 
                \hline
                Argentina & corn & 2017-2023 & 21 & 147 & 1,003,133 \\
                Argentina & soybean & 2017-2023 & 29 & 289 & 2,103,250 \\
                Argentina & wheat & 2017-2022 & 13 & 61 & 497,651 \\
                \hline
                Germany & wheat & 2016-2022 & 6 & 188 & 306,843 \\
                Germany & rapeseed & 2016-2022 & 6 & 111 & 306,843 \\
                \hline
                Uruguay & soybean & 2018-2022 & 10 & 572 & 2,177,206 \\
                \hline
            \end{tabular}
        \end{table}
    
        \begin{figure}
            \centering
            \adjustbox{center=\textwidth}{\includegraphics[width=1.1\linewidth]{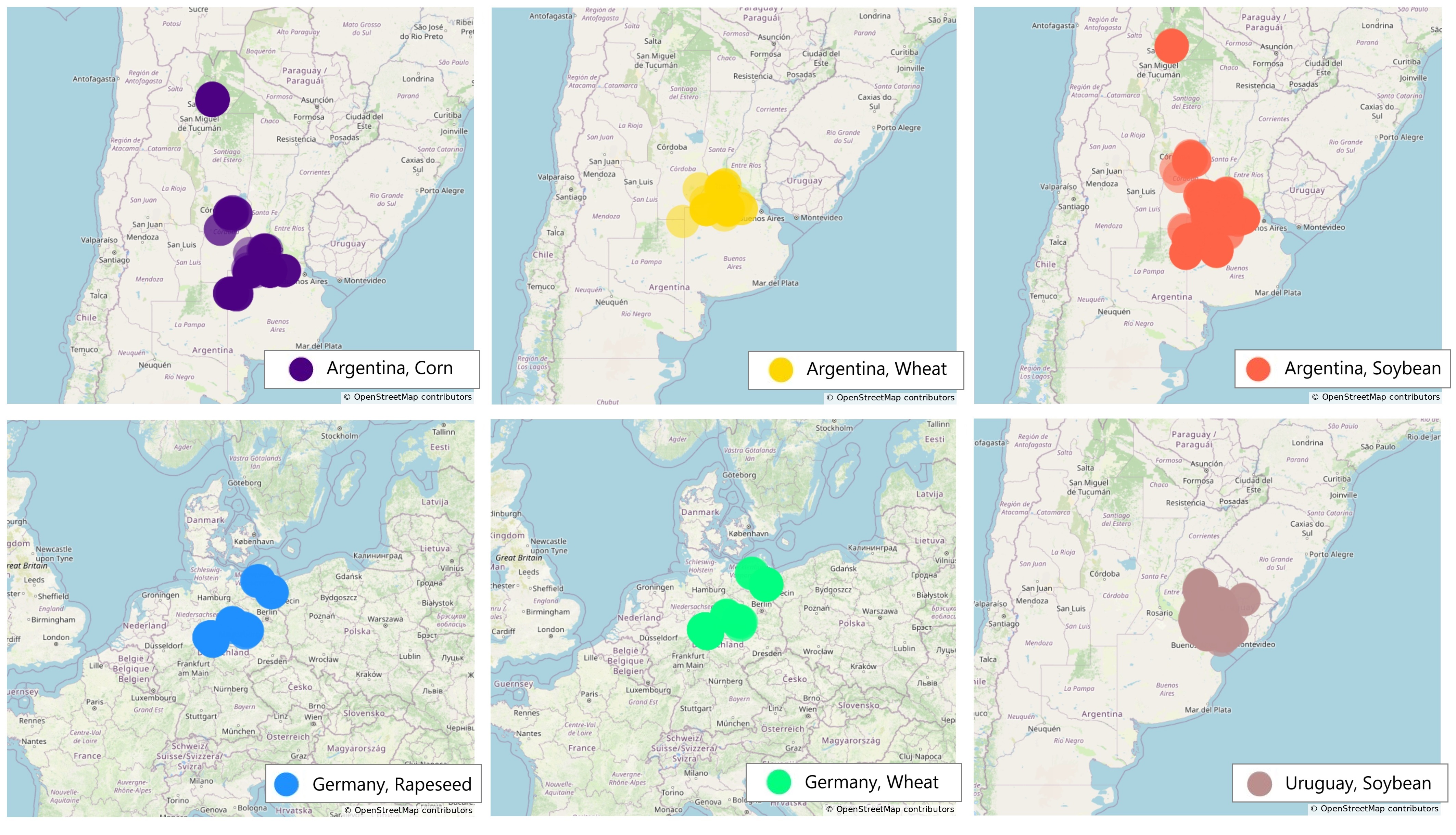}}
            \caption{Geolocation of crop fields in Argentina, Uruguay, and Germany. The buffer sizes are large to protect data confidentiality. Map generated using OpenStreetMap data \citep{osm} and Plotly python package.}
            \label{fig:data_in_map}
        \end{figure}
        
        Yield maps derived from data collected by combine harvesters are used as ground truth, at harvesting date. As the combine traverses the field, it cuts and collects the crop in a single pass, while recording information at equidistant data points at a high spatial resolution. Each point is characterized by various features, including geographic coordinates and the harvested yield in tons per hectare (t/ha). 
        
        To harmonize the raw yield data, we employ a standardized preprocessing pipeline. This includes reprojecting the coordinate reference system, standardizing feature naming conventions, and removing erroneous entries related to position, timestamp, yield, moisture, and non-operational harvesters. Additionally, zero-yield points and agronomically unrealistic values are filtered out. Data points are further refined using statistical thresholds to ensure that yield values remain within three standard deviations. 

        The processed point vector data is subsequently rasterized into 10-meter resolution yield maps, aligned with the corresponding satellite imagery raster data. An overview of the yield datasets is provided in Table \ref{tab:yield_datasets}, while a visual representation of their spatial distribution is shown in Figure \ref{fig:data_in_map}.

    \subsection{Input Modalities}\label{app:input_data}
        Four modalities, all derived from remote sensors, are used for the yield prediction task; Satellite data, weather data, \gls{dem} and soil properties.
        
        \begin{table*}[h]
        \caption{Characteristics of satellite and weather features, with corresponding temporal (Tp.Res.) and spatial (Sp.Res.) resolutions.}
        \centering
        \small
        \begin{tabular}{llccc}
            \hline
            Modality & \textbf{Dynamic features} & \textbf{Source} & \textbf{Sp.Res.} & \textbf{Tp.Res.} \\ \hline
            \multirow{12}{*}{Satellite} & B01 - Coastal Aerosol & S2 & 60 m & 5 days \\
             & B02 - Blue & S2 & 10 m & 5 days \\
             & B03 - Green & S2 & 10 m & 5 days \\
             & B04 - Red & S2 & 10 m & 5 days \\
             & B05 - Red Edge 1 & S2 & 20 m & 5 days \\
             & B06 - Red Edge 2 & S2 & 20 m & 5 days \\
             & B07 - Red Edge 3 & S2 & 20 m & 5 days \\
             & B08 - NIR & S2 & 10 m & 5 days \\
             & B8A - Narrow NIR & S2 & 20 m & 5 days \\
             & B09 - Water vapour & S2 & 60 m & 5 days \\
             & B11 - SWIR 1 & S2 & 20 m & 5 days \\
             & B12 - SWIR 2 & S2 & 20 m & 5 days \\ 
             & Scene Classification Layer & S2 & 20 m & 5 days \\ \hline
            \multirow{4}{*}{Weather} & Max temperature & ERA5 & 30 km & Daily \\
             & Mean temperature & ERA5 & 30 km & Daily \\
             & Min temperature & ERA5 & 30 km & Daily \\
             & Total precipitation & ERA5 & 30 km & Daily \\ \hline
        \end{tabular}
        
        \label{tab:input_data_temp}
        \end{table*}
        \begin{table*}[t]
            \caption{Characteristics of soil and terrain elevation features, with corresponding spatial resolutions (Sp.Res.).}
            \centering
            \small
            \begin{tabular}{llcc}
                \hline
                Modality & \textbf{Static features} & \textbf{Source} & \textbf{Sp.Res.} \\ \hline
                \gls{dem} & Elevation & SRTM & 30 m \\
                 & Slope & SRTM & 30 m \\
                 & Curvature & SRTM & 30 m \\
                 & TWI & SRTM & 30 m \\
                 & Aspect & SRTM & 30 m \\ \hline
                \multirow{8}{*}{Soil} & CEC & SoilGrids & 250 m \\
                 & CFVO & SoilGrids & 250 m \\
                 & Nitrogen & SoilGrids & 250 m \\
                 & pH-H2O & SoilGrids & 250 m \\
                 & Sand & SoilGrids & 250 m \\
                 & Silt & SoilGrids & 250 m \\
                 & SOC & SoilGrids & 250 m \\
                 & Clay & SoilGrids & 250 m \\ \hline
            \end{tabular}
            
            \label{tab:input_data_stat}
        \end{table*}
        
        \begin{itemize}
            
            \item \textbf{Satellite} data originates from the \gls{s2} mission and consists of two sun-synchronous satellites operating in tandem. Their swath width of 290 km ensures a high revisit frequency of five days at the equator.
            All available \gls{l2a} Bottom-of-Atmosphere reflectance images from the seeding to the harvesting date are downloaded from Copernicus \citep{CopernicusSentinel2}.
            The satellite images contain 12 spectral bands (i.e., channels), as listed in Table \ref{tab:input_data_temp}. Each band corresponds to a specific wavelength range and spatial resolution. All bands are upsampled to the highest resolution, i.e., 10x10 meters. 
            In addition to the spectral bands of the satellite image, the \gls{l2a} product includes a \gls{scl}
            at a spatial resolution of 20 meters. This map categorizes pixels into one of 12 predefined classes, including missing data, defective pixels, topographic cast shadows, cloud shadows, vegetation, non-vegetation, water, unclassified, medium-probability clouds, high-probability clouds, thin cirrus, and snow or ice. These classes do not correspond directly to traditional land cover classification layers. In our methodology, we utilize this classification by encoding each pixel’s category as a one-hot vector, which is subsequently combined with the spectral band data.

            \item Hourly \textbf{weather} features were initially retrieved from the \gls{ecmwf} ERA5 global reanalysis dataset \citep{hersbach2020era5} and transformed into daily aggregates. These include minimum, average, and maximum temperatures, as well as total precipitation. 
            Temperatures are calculated for the air at 2 meters above the surface.
            All available observations from the seeding to the harvesting date are collected.
            Weather data, originally available at a spatial resolution of 31x31 km, is aggregated at the field level.
            
            \item \textbf{Soil} data is taken from SoilGrids platform \citep{hengl2017soilgrids250m}, a system for global digital soil mapping using state-of-the-art machine learning to map soil properties across the globe. SoilGrids provides access to soil maps via the Web Coverage Service \citep{WCS}, from which we download maps for eight soil properties: cation exchange capacity (CEC), volumetric fraction of coarse fragments (CFVO), nitrogen content, soil pH (pH-H2O), sand, silt, soil organic carbon (SOC), and clay content, at depths of 0–5, 5–15, and 15–30 cm. The data, originally available at a 250m resolution, is upsampled to a 10m resolution using cubic spline interpolation.
            
            \item \textbf{\gls{dem}} is derived from NASA’s \gls{srtm} \citep{farr2000shuttle}, using U.S. Geological Survey Earth Explorer \citep{usgs}. The data is originally provided at a 30m resolution and upsampled similarly to the soil data. Additional features, including aspect, curvature, slope, and topographic wetness index (TWI), were derived using the RichDEM Python tool \citep{barnes2016richdem}.
            
        \end{itemize}

        Tables \ref{tab:input_data_temp} and \ref{tab:input_data_stat} summarize the features in each input modality, along with their spatial and temporal resolutions. For static features, only the spatial resolution is provided. During the modeling process, all bands are standardized using min-max rescaling, and the missing time steps are padded with -1.

    \subsection{Data preparation}

        \begin{figure}
            \centering
            \includegraphics[width=\linewidth]{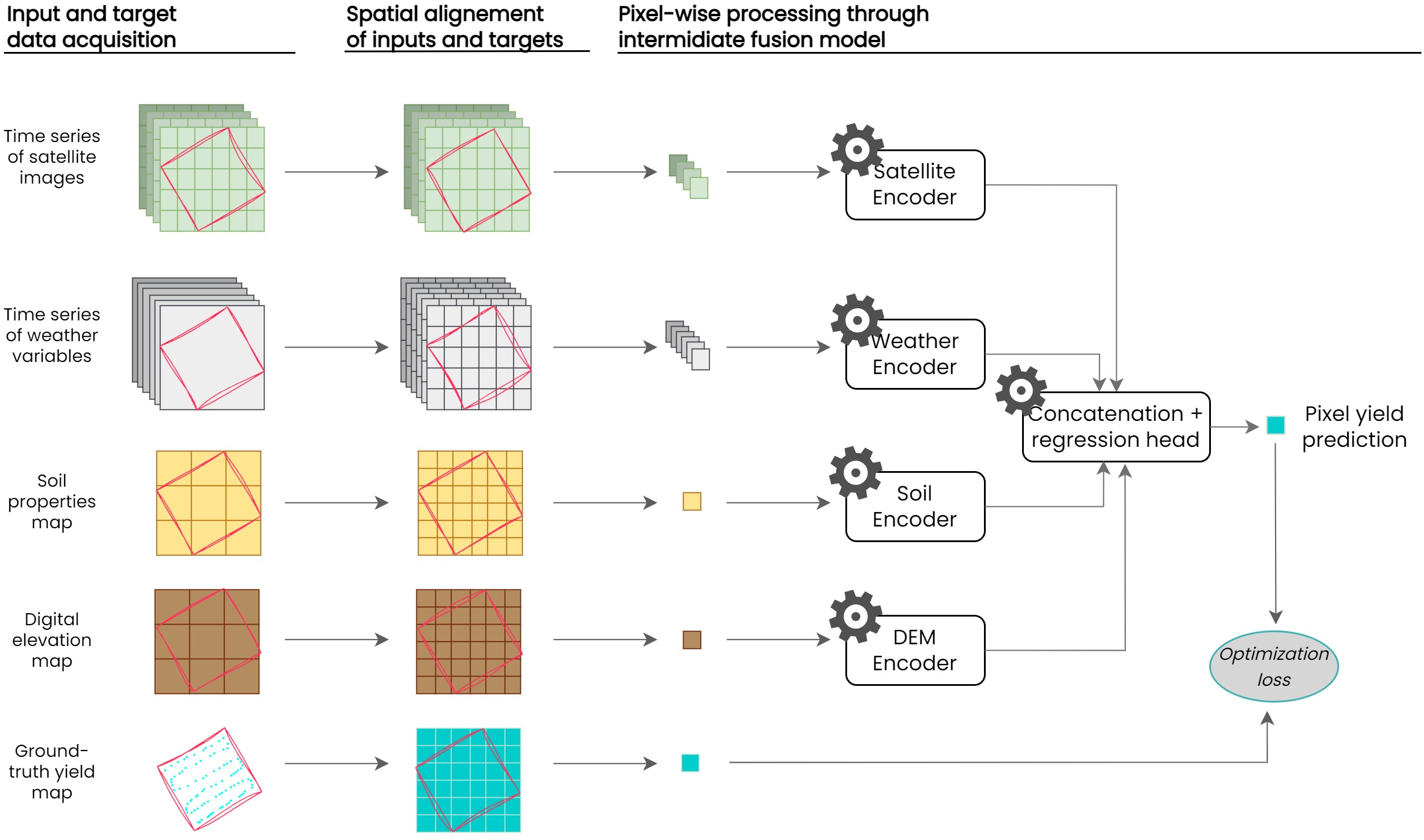}
            \caption{Diagram illustrating the overall workflow, from input modalities and target data acquisition to model processing and pixel-wise crop yield prediction.
            This example shows a pixel taken from a single field (i.e., the red rectangle), while in practice the pixels composing the training data are extracted from multiple fields.}
            \label{fig:overallflow}
        \end{figure}
        
        The different modalities used in this study vary in both spatial and temporal resolution, as detailed in Tables \ref{tab:input_data_temp} and \ref{tab:input_data_stat}. Since each modality is processed separately using an intermediate fusion approach, the temporal resolution of time-dependent modalities (i.e., satellite and weather data) remains unchanged. For technical implementation purposes, we only standardize the sequence lengths across each modality by padding shorter sequences with -1 to account for missing time steps.

        Spatial resolution, on the other hand, is handled differently, as the data is processed on a pixel-wise basis. Thus, we first ensure spatial alignment across all modalities, and we use satellite data as the reference, specifically its high-resolution bands (10m resolution). 
        For low-resolution satellite bands are upsampled using bilinear resampling, while the \gls{scl} mask we use nearest-neighbor resampling. For soil and \gls{dem}, cubic spline interpolation is applied. In the case of weather variables, due to their inherently coarse spatial resolution, they are duplicated across the field. 
        Figure \ref{fig:overallflow} illustrates the data preparation steps for spatially aligning the modalities and target data, as well as the pixel-wise processing using the intermediate fusion model.

        The resulting input sample is a single pixel $\mathbf{X}$, which consists of time series data from satellite bands (including the \gls{scl} classes) $\mathbf{x}^{sa}$ and weather variables $\mathbf{x}^{w}$, along with static soil properties $\mathbf{x}^{so}$ and \gls{dem} data $\mathbf{x}^{dem}$:
        $$
        \mathbf{X} = (\mathbf{x}^{sa},\mathbf{x}^{w},\mathbf{x}^{so},\mathbf{x}^{d}); 
        $$
        $$
        \text{ where: }
        \mathbf{x}^{sa} \in \mathcal{R}^{T_{sa} \times 25}, 
        \mathbf{x}^{w} \in \mathcal{R}^{T_{w} \times 4},
        \mathbf{x}^{so} \in \mathcal{R}^{24},
        \mathbf{x}^{dem} \in \mathcal{R}^{5}.
        $$
        Here, $T_{sa}$ and $T_w$ are the sequence lengths of the satellite and weather time series, respectively, which vary across datasets.
        Note also that, in addition to the 12 satellite bands, $\mathbf{x}_{sa}$ further contains the \gls{scl} classes represented as one-hot encodings. These contain the 12 original classes, along with an additional class to indicate that the time step is padded.

        As described in Section \ref{sec-transformers}, the Transformer models are additionally supplied with two other inputs: the dates of the satellite and weather modalities. These dates serve as positional encodings for the respective sequences, as the model itself does not inherently understand order.
        To compute the positional encodings, we calculate the number of days $n_t=n_{days}$ from the first day of the two calendar years. Each time step $t$ in the sequence is then assigned its corresponding (unique) number. A function is used to generate a positional encoding matrix $\mathbf{P}$ of shape $(T,d)$, where $T \in \{T_{sa},T_{w}\}$. The following sinusoidal functions are used:

        \begin{equation}
        \begin{split}
        \begin{aligned}
        p_{t, j} &= \sin\left(\frac{n_t}{10000^{2j/d}}\right), & \text{for } 0 \leq j < \frac{d}{2}, \
        \\p_{t, j + \frac{d}{2}} &= \cos\left(\frac{n_t}{10000^{2j/d}}\right), & \text{for } 0 \leq j < \frac{d}{2}.
        \end{aligned}
        \end{split}
        \end{equation}
        where $2j$ and $2j+1$ are the embedding dimension indices. Subsequently, the positional encoding matrix $\mathbf{P}$ is added to the input embeddings of shape $(T,d)$ before feeding them to the first Transformer layer.

    \subsection{Data splitting}
        As described in Section \ref{ssec:model_ft}, we split each dataset into training (60\%), validation (20\%) and test (20\%) sets.
        This partitioning aligns with our focus on yield simulation (gap-filling historical records) rather than forecasting future yields.
        Since each input sample represents a pixel from a field, we grouped samples by field before splitting the data, to ensure that the model encounters unseen fields in the validation and test splits. To maintain a consistent data distribution, we stratified the splits by year, ensuring that each split contains data from all years.

\section{Experiments and Results}\label{sec:results}

    In this section, we report the results of all conducted experiments, beginning with 
        model evaluation at subsection \ref{ssec:model_eval},
        followed by the analysis of intermediate representations and temporal attributions in subsections \ref{ssec:prob_repr}, \ref{ssec:temp_attr}, and \ref{ssec:weather_events}, 
        before presenting modality importance estimation techniques and results in \ref{ssec:mod_imp}.

    \subsection{Model evaluation}\label{ssec:model_eval}

        Here we provide answers to \ref{rq1}, which goal is to identify the best performing multimodal network architecture for yield prediction, and to select the model to explain when addressing \ref{rq2}-\ref{rq5}.

        \subsubsection{Quantitative results} 
            To assess the performance of the models described in Section \ref{sec:modeling}, we use the validation set to compare multiple instances, according to their \gls{r2} score. We subsequently select the best-performing models per architecture, for which we report the scores in Table \ref{tab:models_res}, including the \gls{mae} and \gls{rmse} metrics. 
            The complete architecture of each model is described in \ref{app:model_ft2}.
            The subfield-level scores refer to the average performance across the pixel samples, while the field-level refers to the accuracy of predicting the average yield per field, by averaging the pixel-level target values and predictions.

            We first observe that the subfield-level metrics are consistently worse than the field-level values. This difference arises from the increased complexity of the task of accurately predicting the yield values for individual pixels, as compared to estimating the average yield. The pixel-level predictions demand a higher level of detail and precision, making it a more challenging task than field-level predictions, where averaging helps to smooth out local variations.

            Comparing model architectures, the Transformer model demonstrates a clear advantage in performance at the subfield-level, compared to the convolutional and recurrent networks. The scores become closer at the field-level, where the Transformer remains optimal, followed closely by the \gls{1d-cnn}. This observation aligns with the efficient performance of the temporal convolutional networks achieved in other studies \citep{najjar2024data}. We also notice that at the field-level, the attention mechanism improves the performance of the recurrent network, highlighting the role of attention in enhancing the predictive accuracy in \gls{ml}.

            We further compare the inference time of the best performing models in Table \ref{tab:models_inf_time}. Inference experiments were conducted over a batch of 1000 samples, and results were averaged across 500 batches. The machine setup included an NVIDIA V100 GPU with 16GB of memory, 50 CPUs, CUDA version 11.8, Python version 3.8.10 and PyTorch version 1.14.0.
            We observe in Table \ref{tab:models_inf_time} that without a GPU, the convolutional model exhibits the highest inference speeds, slightly exceeding half a second per batch, on average, followed by the Transformer model which slightly exceeds two seconds. 
            In contrast, recurrent networks demonstrate the slowest speeds, primarily due to their sequential processing of time steps. 
            When using a GPU, the speed ranking is reversed, as the recurrent networks achieve the fastest processing times. Nevertheless, the inference times of all models remain mostly below 22 milliseconds.
            
            \begin{table*}[t]
                \caption{Comparison of model performance evaluated on the subfield-level (i.e. pixel level) and the field-level on the test set.}
                \centering
                \small
                \adjustbox{center=\textwidth}{\begin{tabular}{cclccclccc}
                    \hline
                    Model & \multicolumn{1}{l}{\# Parameters} &  & \multicolumn{3}{c}{Subfield-Level} &  & \multicolumn{3}{c}{Field-Level} \\ \cline{4-6} \cline{8-10} 
                     & \multicolumn{1}{l}{} &  & R² & RMSE (t/ha) & MAE (t/ha) &  & R² & RMSE (t/ha) & MAE (t/ha) \\ \cline{1-2} \cline{4-6} \cline{8-10} 
                    1D-CNN & 6,333,377 &  & 0.42 & 2.58 & 1.94 &  & 0.74 & \textbf{1.41} & \textbf{1.02} \\
                    LSTM & 54,977 &  & 0.41 & 2.61 & 2.01 &  & 0.71 & 1.71 & 1.32 \\
                    ALSTM & 38,017 &  & 0.41 & 2.59 & 1.99 &  & 0.74 & 1.47 & 1.19 \\
                    Transformer & 147,073 &  & \textbf{0.52} & \textbf{2.35} & \textbf{1.74} &  & \textbf{0.78} & 1.42 & \textbf{1.02} \\ \hline
                \end{tabular}}
                \label{tab:models_res}
            \end{table*}
            
            \begin{table}[t]
                \centering
                \caption{Comparison of model inference time in seconds.}
                \label{tab:models_inf_time}
                \small
                \begin{tabular}{clcclcc}
                    \hline
                    Model       &  & \multicolumn{2}{c}{CPU} &  & \multicolumn{2}{c}{GPU}           \\ \cline{3-4} \cline{6-7} 
                                &  & Mean (s)       & STD (s)        &  & Mean (s)  & STD (s)                       \\ \cline{1-1}\cline{3-4} \cline{6-7} 
                    1D-CNN      &  & 0.549      & 0.090      &  & 0.006 & 0.013                     \\
                    LSTM        &  & 4.048      & 0.575      &  & 0.003 & 0.002                     \\
                    ALSTM       &  & 2.800      & 0.482      &  & 0.003 & 0.004 \\
                    Transformer &  & 2.055      & 0.156      &  & 0.020 & 0.002                     \\ \hline
                \end{tabular}
            \end{table}

        \subsubsection{Transformer configuration}
            
            To investigate the behavior of different configurations of the Transformer-based model, we report the evaluation metrics on the validation and test sets across various model setups, focusing on the five best-performing instances, as shown in Table \ref{tab:Trans_models_res_sbfld}.
            We observe that the models achieve comparable performance on the validation set: 
                the \gls{r2} scores range between 0.75 and 0.77, 
                the \gls{mae} falls within 1.35 and 1.43 t/ha, 
                while the \gls{rmse} values are within the range of 1.85-1.92 t/ha.
            However, large variance is observed in the test set results. The range of \gls{r2} values varies between 0.39 and 0.52, while \gls{mae} ranges between 1.74 and 1.97 t/ha, and \gls{rmse} ranges from 2.35 to 2.64 t/ha.
            The models ranked first and fifth achieve the best performance, with a moderate margin above other models.

            These observations suggest that the generalization capacity of the model on the validation set does not necessarily transfer to the test set. The similar performance observed in the validation set further indicates that changes in model parameters, such as the number of heads or layers, have a relatively minor impact on overall performance.
            
            Based on these results, and in order to prioritize simplicity and ease of interpretability over marginal gains in evaluation metrics, the explanation experiments in the following subsections will focus on the fifth model, which we will refer to as the ``\textit{selected}" model. Achieving the closest performance to the best model on the test sets, the selected model offers the advantage of using a single head in the Transformer encoders. 
            This choice simplifies the estimation and interpretation of the \gls{ar} and \gls{ga} explanation results, as it avoids the need to interpret multiple heads individually or average the results across heads \citep{abnar2020quantifying,chefer2021generic}. While interpreting individual heads increases the complexity of model interpretability, averaging the results risks suppressing patterns learned exclusively by each head \citep{voita2019analyzing}.

            \begin{table*}[]
                \caption{Comparison of Transformer models performance on the subfield-level. Best and second-best scores are highlighted in bold and underlined, respectively.}
                \label{tab:Trans_models_res_sbfld}
                \centering
                \small
                \adjustbox{center=\textwidth}{\begin{tabular}{ccccccccccc}
                    \hline
                    \multirow{2}{*}{\begin{tabular}[c]{@{}c@{}}Hidden\\ size\end{tabular}} & \multirow{2}{*}{Heads} & \multirow{2}{*}{Layers} &  & \multicolumn{3}{c}{Validation set} &  & \multicolumn{3}{c}{Test set} \\ \cline{5-7} \cline{9-11} 
                     &  &  &  & R² & RMSE (t/ha) & MAE (t/ha) &  & R² & RMSE (t/ha) & MAE (t/ha)\\ \cline{1-3} \cline{5-7} \cline{9-11}
                    64 & 4 & 2 &  & \textbf{0.77} & \textbf{1.85} & \textbf{1.35} &  & \textbf{0.52} & \textbf{2.35} & \textbf{1.74} \\
                    32 & 2 & 2 &  & \uline{0.76} & \uline{1.86} & \uline{1.36} &  & 0.39 & 2.64 & 1.97 \\
                    64 & 1 & 2 &  & 0.75 & 1.91 & 1.38 &  & 0.40 & 2.61 & 1.95 \\
                    32 & 2 & 6 &  & 0.75 & 1.91 & 1.42 &  & 0.44 & 2.54 & 1.93 \\
                    32 & 1 & 4 &  & 0.75 & 1.92 & 1.43 &  & \uline{0.48} & \uline{2.44} & \uline{1.83} \\[0.1em] \hline
                \end{tabular}}
            \end{table*}

        \subsubsection{Performance across years and farms}
            Since our focus was on historical yield simulation, we conducted the above model optimization using standard training/validation/test splits. Other prediction scenarios, including \gls{lofo} and \gls{loyo}, might require different data splitting approaches.
            While neither our data collection nor model fine-tuning were optimized for these schemes, we nevertheless trained and evaluated our selected architecture using such splits to provide readers with insights into the dataset and model's potential and limitations for these applications. These results are included in \ref{app:lofo_loyo}.
             
        \subsubsection{Qualitative results}
\begin{figure*}[t]
                \centering  
                \includegraphics[width=\linewidth]{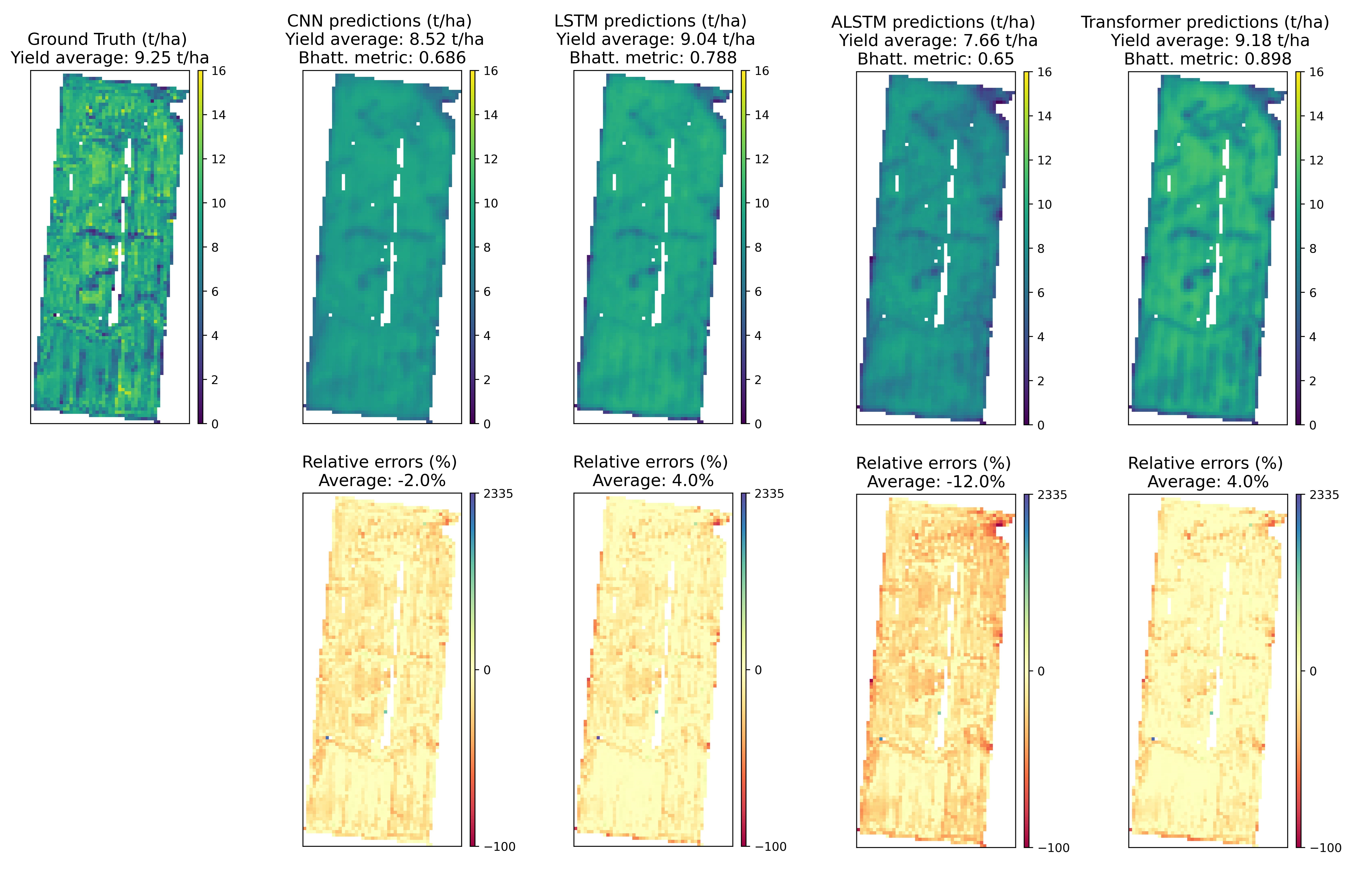} 
                \caption{Qualitative results on Field-A. From left to right: ground-truth and predicted yield values of the \gls{1d-cnn}, \gls{lstm}, \gls{alstm}, and Transformer models, with the relative prediction error displayed at the bottom.}
                \label{fig:pred_map_good_field}
            \end{figure*}

\begin{figure*}[t]
                \centering  
                \includegraphics[width=1.05\linewidth]{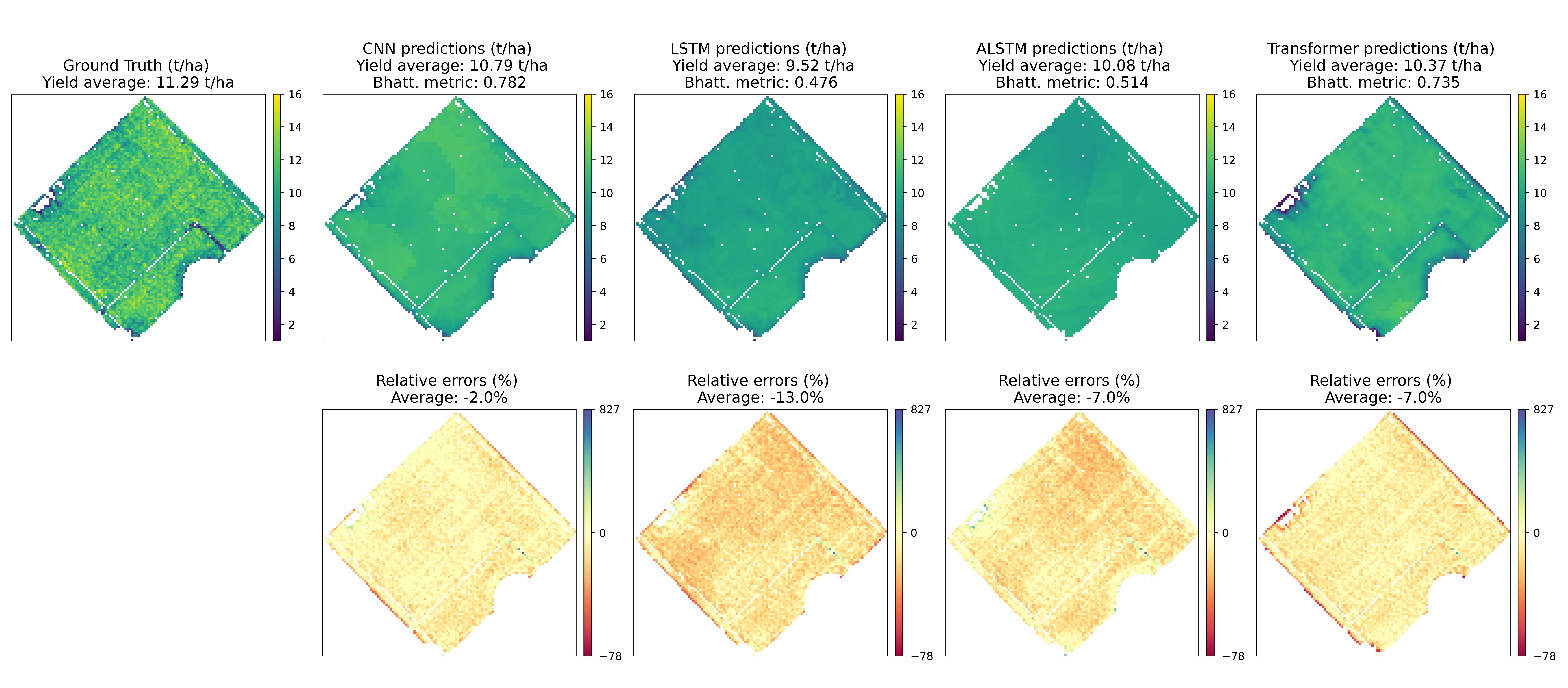}
                \caption{Qualitative results on Field-B. From left to right: ground-truth and predicted yield values of the \gls{1d-cnn}, \gls{lstm}, \gls{alstm}, and Transformer models, with the relative prediction error displayed at the bottom.}
                \label{fig:pred_map_bad_field}
            \end{figure*}

            To visually compare the performance of the selected model from each architecture type, we selected two corn fields from the validation set, referred to as Field-A and Field-B, and plot in Figures \ref{fig:pred_map_good_field} and \ref{fig:pred_map_bad_field} the corresponding target, prediction, and relative error maps. 
            These visualizations help illustrate how well the models capture spatial patterns in yield predictions and highlight areas where they may struggle. 
            The fields were chosen based on the evaluation of the selected Transformer model:  Field-A, where the model performed well, and Field-B, where the model performed poorly.

            The results for Field-A are shown in Figure~\ref{fig:pred_map_good_field}. 
            The top row displays the target yield values alongside the predicted values from the best-performing \gls{1d-cnn} first, followed by the \gls{lstm}, the \gls{alstm}, and finally the selected Transformer model. The average yield is indicated above each map, alongside the Bhattacharyya metric scores, which measures the captured yield variance. The second row shows the corresponding error maps for each model.
            We notice that the Transformer model provides the most accurate approximation of the average yield, with 9.18 t/ha compared to the ground truth of 9.25 t/ha, in addition to capturing the highest yield variance, as indicated by its high Bhattacharyya score of 89.8\%. 
            The \gls{lstm} model follows closely, achieving the same average relative error of 4\% as the Transformer model.
            In contrast, both the \gls{1d-cnn} and \gls{alstm} model struggle to capture the in-field yield variance, as they achieve lower Bhattacharyya scores of 68.6\% and 65\%, respectively. 
            The \gls{alstm} model particularly underestimates the yield at several pixels and regions of the field, as shown in the corresponding relative error map.

            Figure \ref{fig:pred_map_bad_field} presents the results for Field-B, where the Transformer model did not perform very well.
            The \gls{1d-cnn} and Transformer models demonstrate a relatively higher ability to approximate the average yield and capture its variance, with the convolutional model slightly outperforming the Transformer model. In contrast, the \gls{lstm} and \gls{alstm} models achieve lower accuracy in predicting the yield, on average, and tend to underestimate the yield across many pixels, as depicted in the respective error maps.
            Interestingly, most models face challenges in accurately predicting the yield near the field borders, with the \gls{alstm} model being an exception, showing better performance in these regions.
            
            Overall, despite the challenges in Field-B, the Transformer model maintains comparatively good performance relative to the other models. Considering CPU and GPU inference times, performance improvements, and interpretability, we believe the Transformer model offers a well-balanced choice for the subsequent interpretability analysis.

    \subsection{Learned Representations}\label{ssec:prob_repr}

        In this section, we address \ref{rq2} and evaluate the information content of intermediate model representations using linear probing, focusing on the selected Transformer-based model. 
        Next, we analyze the attention weight matrices learned by the model, evaluating their similarity across pixels within the same field, and examining how these weights are distributed across the different layers of the Transformer encoders.

        \subsubsection{Linear Probing}

            \begin{figure}
              \centering
              \includegraphics[width=0.6\linewidth]{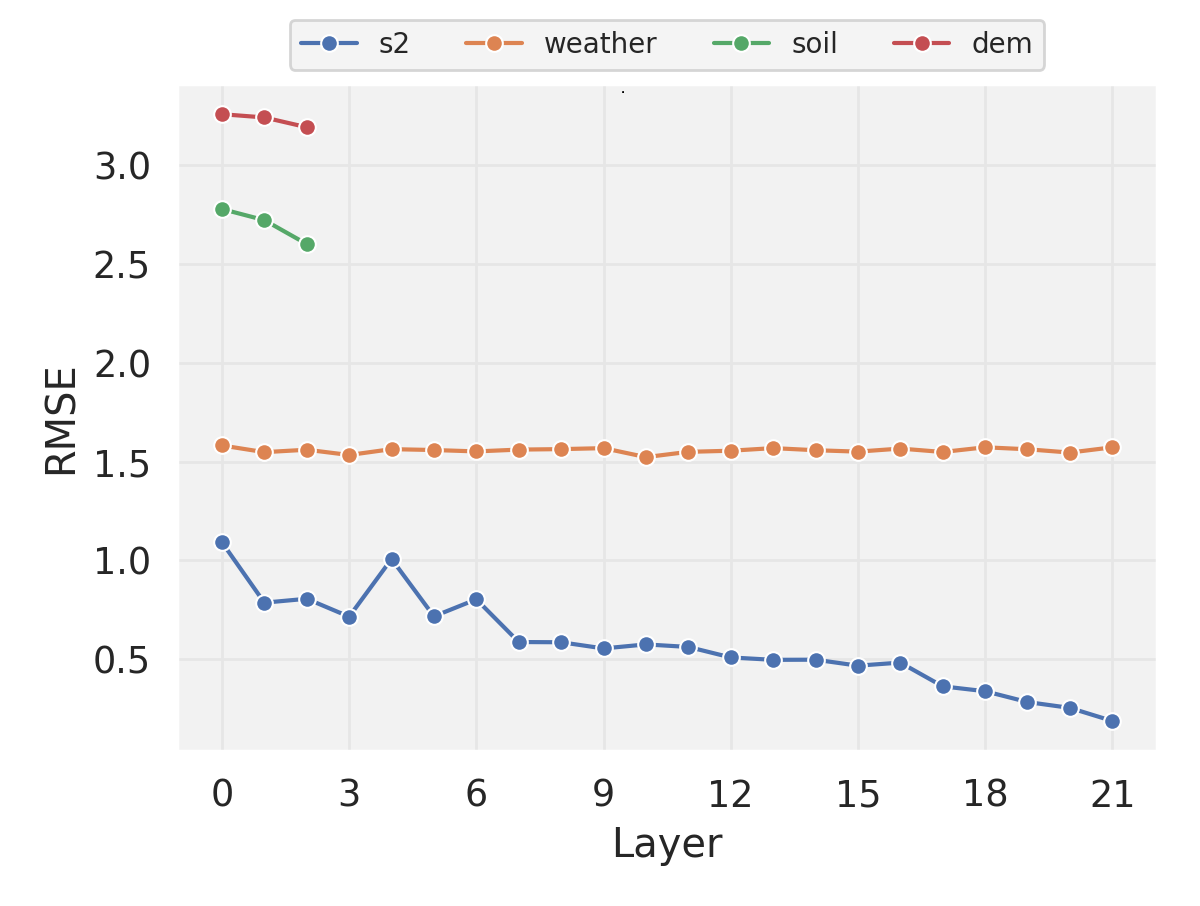} 
              \caption{\gls{rmse} (t/ha) test scores of the linear probes attached to the modality encoders.}
              \label{fig:lp_rmse}
            \end{figure}

            We investigate the linear separability of the intermediate layers of the selected Transformer model, as described in Section \ref{ssec:lp}. 
            To facilitate this analysis, we randomly select 100,000 samples, representing approximately 10\% of the corn dataset, using 90\% of these samples to train linear probes and the remaining 10\% for testing. 
            For each layer, we compute its output given the selected samples as inputs, flatten these latent representations, and then use them to train a linear model to predict the model's final yield prediction.
            The \gls{rmse} scores on the test set are presented in Figure~\ref{fig:lp_rmse}.

            We observe that the intermediate representations learned for the \gls{s2} satellite data demonstrate the highest linear correlation to the predicted values across all layers, followed by the weather data. In contrast, soil and \gls{dem} data show significantly lower linear separability.
            Given the static nature of these two modalities, they were processed using shallow \glspl{mlp}, and they also have low spatial resolution, which contributes to their limited potential to predict the yield. 
            When comparing the temporal modalities, i.e. satellite and weather data, the results indicate that the linear separability of weather data remains nearly constant throughout the Transformer layers, whereas a significant increase is observed across the satellite encoder layers. This trend can be attributed to the higher complexity of the satellite time series, which has the highest spatial resolution and comprises 12 spectral bands, in contrast to the four weather properties used.

        \subsubsection{Attention weights: In-field distribution}\label{ssec:s2_scatterplot}      
        
            Considering that for pixels within the same field yield variations are expected to be minimal and growth conditions are similar, we quantify the similarity of attention weights at the field-level to later aggregate the attention-based explanations at this level.  
            We also examine the correlation of attention weights similarity with the prediction similarity of each pair of pixels.
            This analysis is conducted through the following steps: First, 200 pixels are randomly selected from each field. 
            Then, for each pair of pixels we calculate (i) the cosine similarity between attention weights and (ii) the difference in predicted yield, separately in each field. 
            Finally, scatter plots are generated, where the similarity values are plotted per field and colored according to the corresponding difference in predicted yield.

            \begin{figure}[t]
                \centering  
                \adjustbox{center=\textwidth}{\includegraphics[width=1.1\linewidth]{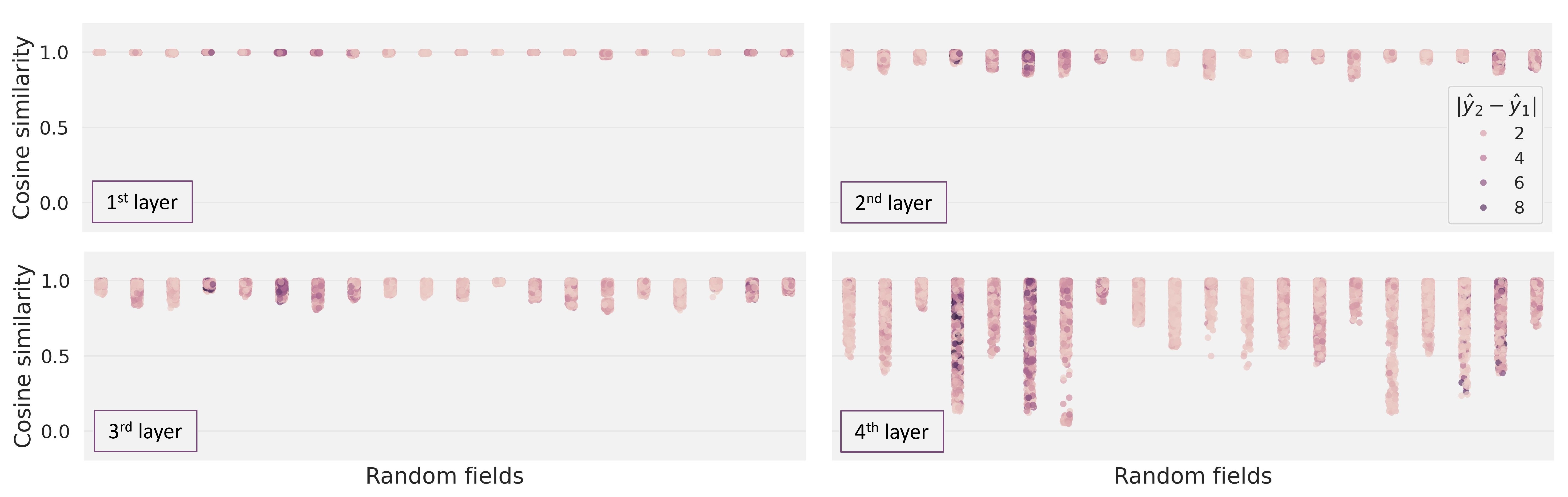}}
                \caption{
                    Cosine similarity of the attention weights from the satellite encoder of multiple pairs of pixels in a consistent set of 20 random corn fields, and the corresponding difference in prediction.  
                }
                \label{fig:field_sim_s2_raw_att}
            \end{figure}
            
            \begin{figure}[t]
                \centering  
                \adjustbox{center=\textwidth}{\includegraphics[width=1.1\linewidth]{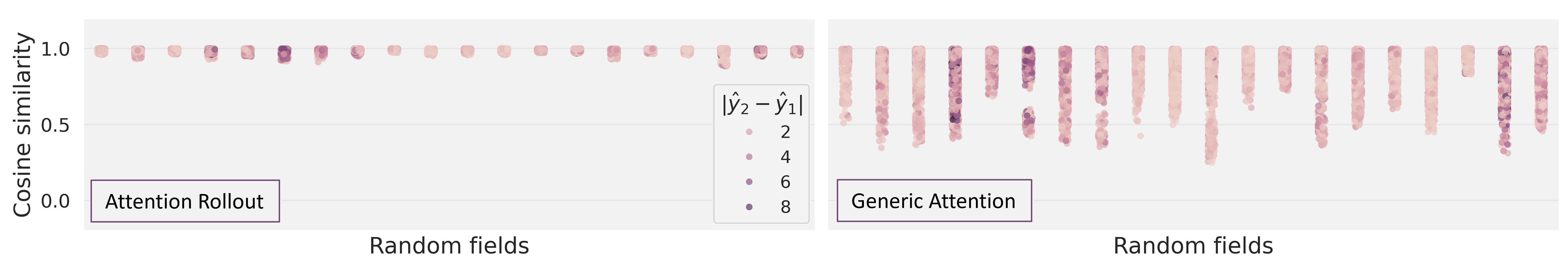}}
                \caption{
                    Cosine similarity of the \gls{ar} and \gls{ga} of the satellite encoder of multiple pairs of pixels in a consistent set of 20 random corn fields, and the corresponding difference in prediction.
                }
                \label{fig:field_sim_s2_roll_gen_att}
            \end{figure}
            
            \begin{table}[t]
                \caption{Minimum, mean, and maximum Spearman correlation values between pairwise cosine similarity scores and prediction differences across 20 corn fields.}
                \centering
                \small
                
                \begin{tabular}{ccccccc}
                    \cline{2-7}
                     & \multicolumn{4}{c}{\textbf{Attention Weights Layers}} & \multirow{2}{*}{\textbf{AR}} & \multirow{2}{*}{\textbf{GA}} \\ \cline{2-5}
                     & \textbf{1st} & \textbf{2nd} & \textbf{3rd} & \textbf{4th} &  &  \\ \cline{1-7} 
                    Min & -0.13 & -0.17 & -0.10 & -0.18 & -0.14 & -0.14 \\
                    Mean & 0.00 & -0.01 & 0.00 & 0.00 & -0.01 & 0.00 \\
                    Max & 0.09 & 0.09 & 0.09 & 0.13 & 0.13 & 0.12 \\ \hline
                \end{tabular}
                
                \label{tab:spear}
            \end{table}
            
            An example in Figure~\ref{fig:field_sim_s2_raw_att} illustrates the results from each layer of the satellite Transformer encoder from 20 random corn fields.
            For the first three layers, the distance between the flattened full attention weight matrices is compared, whereas for the final layer, only the weights attending to the regression token are considered. 
            We notice a pronounced similarity in the three first layers, but it diminishes significantly in the fourth layer in most fields. 
            Additionally, no correlation is visually identified between the absolute prediction error and the distance between the attention weights of the compared pixel pairs. We verify quantitatively the weak correlation using the Spearman correlation metric, as reported in Table \ref{tab:spear}. The results imply that similar predictions are not necessarily associated with a similar distribution of attention across different time steps, even for pixels within the same field.
            
            We also conducted the same analysis to compare the \gls{ar} and \gls{ga} results. 
            As shown in Figure~\ref{fig:field_sim_s2_roll_gen_att}, a strong similarity is noted between the \gls{ar} attributions at the field-level, in contrast to larger differences observed in the \gls{ga} results. 
            We believe that the high similarities observed in the first three layers in Figure \ref{fig:field_sim_s2_raw_att} should not be entirely outweighed by the decreasing similarities in the last layers, which suggests a higher robustness of \gls{ar} compared to \gls{ga}.
            Additionally, a desirable property of attribution methods is low sensitivity, meaning that minor variations in input feature values should not lead to significant changes in the attributions \citep{yeh2019fidelity}. Since pixels from the same field typically experience similar environmental conditions, their input values are expected to be comparable, and consequently, their attributions should exhibit consistency as well.
            Furthermore, the inclusion of gradients in the computation of \gls{ga} could contribute to its high sensitivity, as shown by \citet{ghorbani2019interpretation} in other gradient-based attribution methods. 
            
            For the weather Transformer encoder, we observe perfect similarity across all evaluated fields, irrespective of the method used. This is attributed to the low spatial resolution of weather data, leading to identical input weather values for all pixels within the same field.

        \subsubsection{Attention weights: Layer-wise distribution}\label{par:att_layers}
            
\begin{figure*}[!t]
                    \centering
                    \adjustbox{center=\textwidth}{\includegraphics[width=1.3\textwidth]{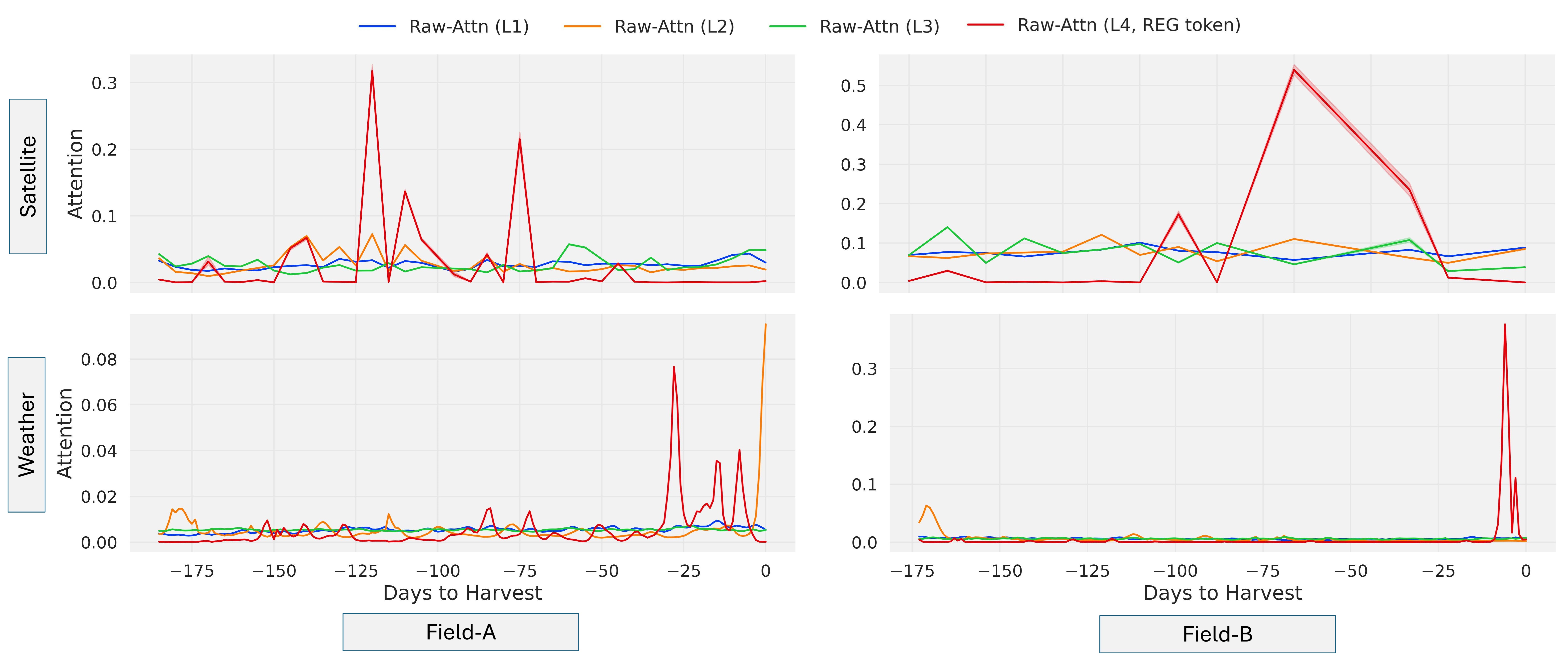}}
                    \caption{
                        Total attention weights attending at each time step for the first 3 attention layers, and the regression token weights in the final layer. 
                        The results are averaged across 200 randomly selected pixels from Field-A, at the top, and Field-B, at the bottom, and are displayed for the satellite (a) and weather (b) Transformer encoders. 
                        The light buffer regions represent the 95\% confidence interval around the average value.
                        }
                    \label{fig:att_layers_s2_weather}
                \end{figure*}

            After assessing the similarity of the attention weights across different pixels, we now study their temporal distribution across different layers. 
            We compute the time series $S^l$ of temporal attentions for each layer $l$ as described in Section \ref{ssec:attn}.
            Figure~\ref{fig:att_layers_s2_weather} presents these results for the temporal modalities, with Field-A shown in the top row and Field-B in the bottom row.
            Results from other fields are displayed in \ref{app:att_layers_rdm_fields}.

                In the case of the satellite time series, we observe that the attention weights from the first layer is distributed smoothly across the entire time series. 
                In contrast, the second and third layers show more peaks, which become significantly more pronounced in the fourth layer.
                These differences across layers were also observed in similar previous studies \citep{xu2021towards}. 
                Moreover, the varying patterns of attention distribution across different fields confirm that each layer is capturing unique temporal dynamics relevant to the conditions of each field.
                For the weather encoder, the attention distribution results reveal that the second and fourth layers exhibit a particularly discriminative behavior across different time steps.

            To understand the information content within temporal attentions, we use Shannon entropy as described in Section \ref{ssec:attn}. The results are presented in Table \ref{tab:attn_entropy} for Field-A, Field-B, and three randomly selected fields (the same fields used in \ref{app:att_layers_rdm_fields}).
                We observe that the first and last layers of the satellite encoder have the lowest entropy values, which indicates that most of the attention mass is concentrated on a few time steps. For the Weather encoder, the lowest scores are observed at the first and third layers.        
                
\begin{figure*}[t]
                    \centering
                    \includegraphics[width=\textwidth]{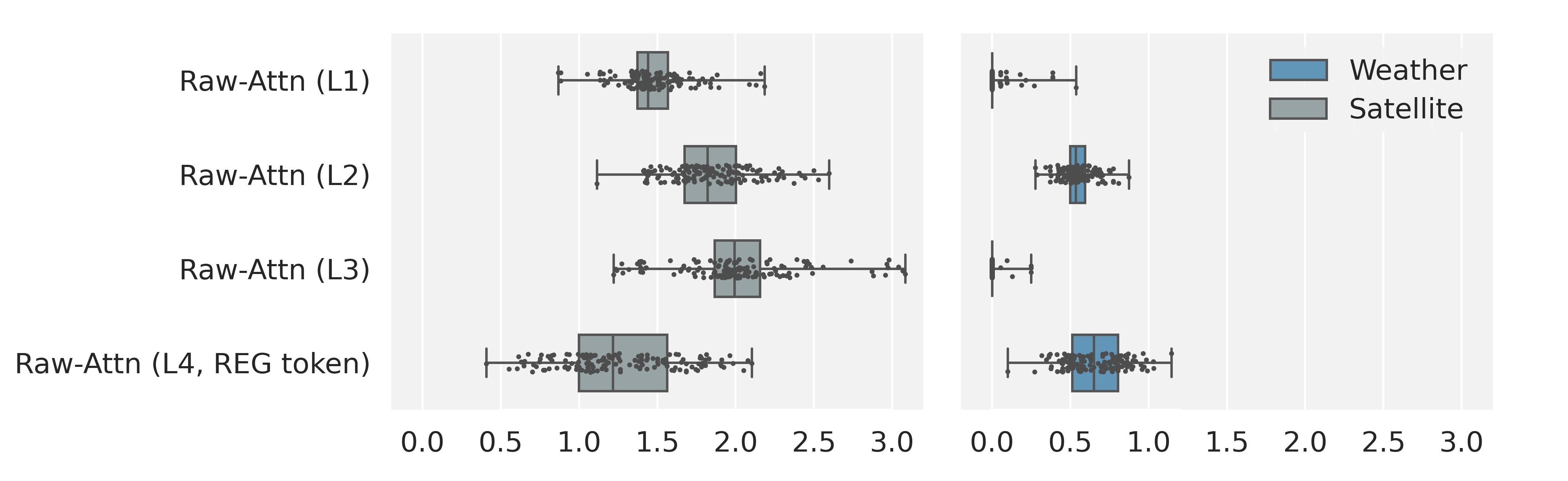}
                    \caption{
                        Shannon entropy of the satellite and weather temporal attentions, averaged across 200 pixels from each corn field.
                        }
                    \label{fig:attn_entropy}
                \end{figure*}

                To gain a broader understanding of the entropy distribution across the corn dataset, Figure \ref{fig:attn_entropy} visualizes entropy scores across all corn fields. The figure confirms the general observation that the lowest entropy scores occur in the first and last layers of the satellite encoder, and in the first and third layers of the weather encoder.
                Additionally, a comparison between the two encoders reveals that the entropy values in the weather encoder layers are generally lower than those in the satellite encoder layers. This suggests that weather information relevant for model prediction is concentrated within fewer time steps, whereas the useful satellite information is distributed more evenly throughout the growth period.


                \begin{table}[t]
                    
                    \caption{Entropy of temporal attentions retrieved from the satellite and weather encoders, averaged across 200 pixels from each field. Lowest entropies per field and modality are highlighted.}
                    \centering
                    \small
                    \begin{tabular}{ccccccc}
                        \cline{3-7}
                        \multicolumn{1}{l}{} &  & \textbf{Field-A} & \textbf{Field-B} & \textbf{R.Field-1} & \textbf{R.Field-2} & \textbf{R.Field-3} \\ \hline
                        \multirow{4}{*}{\begin{tabular}[c]{@{}c@{}}Satellite\\ encoder\end{tabular}} & 1st layer & 1.58 & 2.16 & \textbf{0.88} & \textbf{1.48} & \textbf{1.34} \\
                         & 2nd layer & 2.11 & 2.50 & 1.97 & 1.72 & 1.86 \\
                         & 3rd layer & 2.01 & 3.04 & 1.65 & 2.03 & 2.09 \\
                         & 4th layer & \textbf{1.54} & \textbf{1.77} & 1.50 & 1.50 & 1.68 \\ \hline
                        \multirow{4}{*}{\begin{tabular}[c]{@{}c@{}}Weather\\ encoder\end{tabular}} & 1st layer & \textbf{0.00} & \textbf{0.00} & 0.39 & \textbf{0.00} & \textbf{0.00} \\
                         & 2nd layer & 0.38 & 0.57 & 0.77 & 0.53 & 0.65 \\
                         & 3rd layer & \textbf{0.00} & \textbf{0.00} & \textbf{0.25} & \textbf{0.00} & \textbf{0.00} \\
                         & 4th layer & 0.77 & 0.34 & 0.86 & 0.72 & 0.63 \\ \hline
                    \end{tabular}
                    
                    \label{tab:attn_entropy}
                \end{table}

            These findings highlight the differential use of attention mechanisms across modalities and how different layers of the Transformer model specialize in capturing various temporal aspects of the data, providing insights into how the model interprets and prioritizes different parts of the time series for yield prediction.

    \subsection{Temporal Attributions}\label{ssec:temp_attr}

            \begin{figure*}[t]
                \centering  
                \adjustbox{center=\textwidth}{\includegraphics[width=1.2\linewidth]{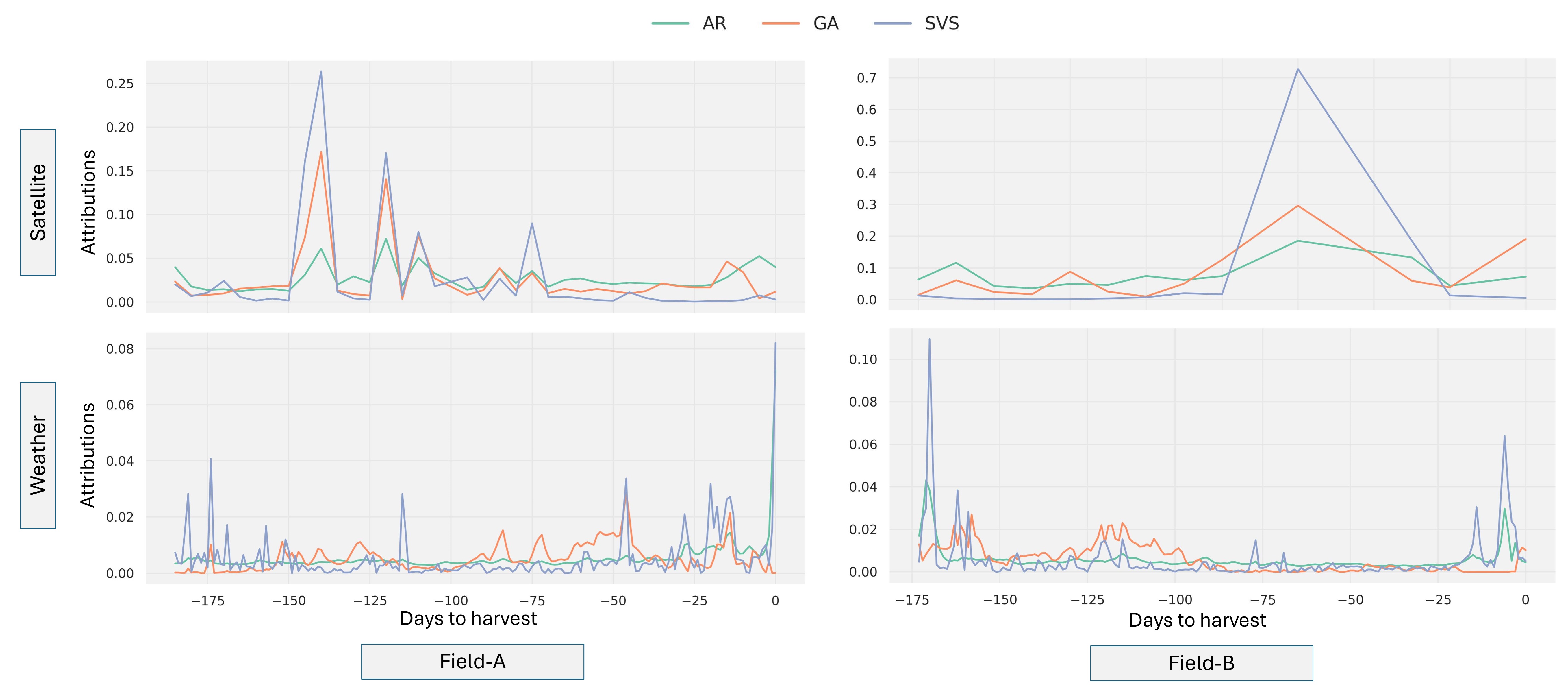}}
                \caption{Field-level average attributions of the satellite and weather modalities, for Fields A and B.
                Due to the high computational cost associated with the \gls{svs} method, we limited the number of pixels sampled per field to 32 pixels. 
                }
                \label{fig:field_temp_attr}
            \end{figure*}         

            We address in this section \ref{rq3} and analyze the temporal attributions provided by the \gls{ar} and \gls{ga} methods. We further include the \gls{svs} method, to compare against post-hoc, model-agnostic estimations of the temporal attributions.
            Figure \ref{fig:field_temp_attr} displays the average attributions for Field-A and Field-B, while results for additional corn fields are provided in \ref{app:temp_attr_rdm_corn}. 
                
\begin{figure*}[t]
                    \centering
                    \includegraphics[width=0.8\textwidth]{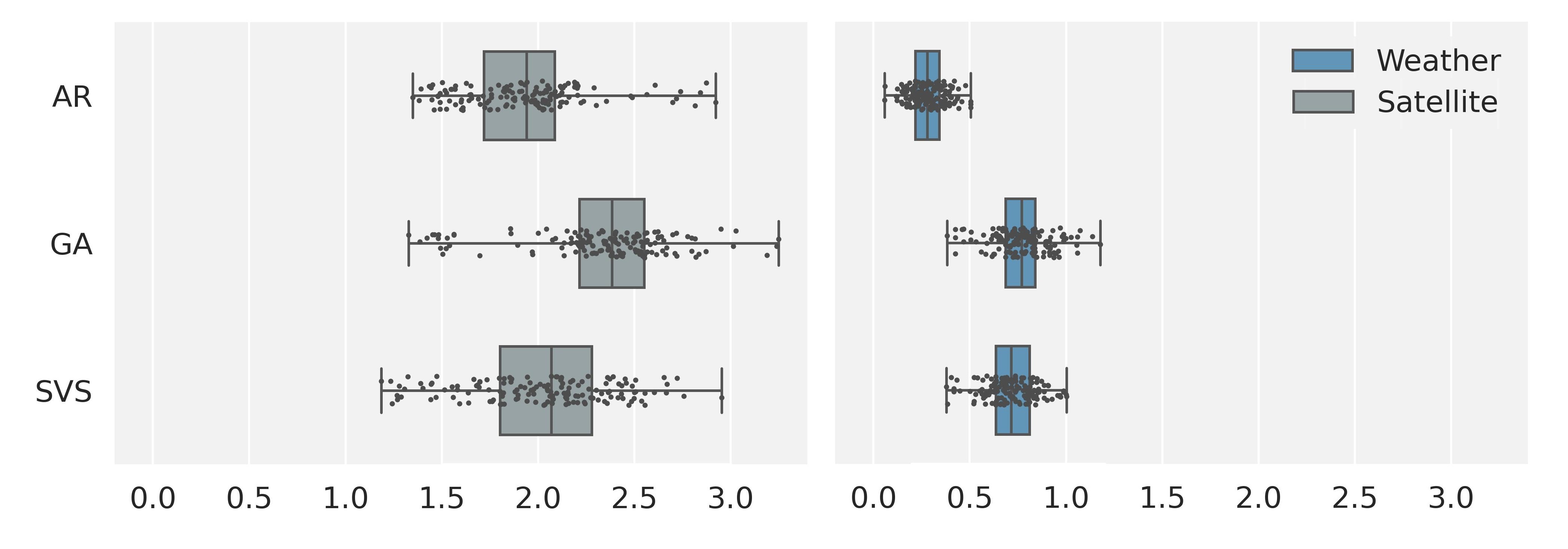}
                    \caption{Shannon entropy of the satellite and weather temporal attributions, averaged across 32 pixels from each corn field.}
                    \label{fig:attr_entropy}
                \end{figure*}

            \paragraph{Entropy Analysis}
                We repeat the entropy analysis to quantify the information compression within the temporal attributions of each modality. The results, shown in Figure \ref{fig:attr_entropy}, reveal that the satellite data attributions exhibit higher entropy compared to those of the weather data. This indicates that the important information in the satellite modality is distributed along a wider range of instances.
                Interestingly, this observation aligns with the findings from the entropy analysis of the attention weights, suggesting that the entropy patterns of the attention mechanism are preserved in the estimated feature attributions.
                Comparing the three methods, \gls{ar} demonstrates the lowest entropy scores, whereas \gls{ga} exhibits the highest scores on average across both modalities. This indicates that \gls{ar} focuses on a smaller subset of instances with significant influence on the model predictions. In contrast, \gls{ga} identifies a broader set of important time steps.  The behavior of \gls{svs} falls between these two patterns.

\begin{figure}[t]
                    \centering  
                    \begin{minipage}{0.47\textwidth}
                        \includegraphics[width=\linewidth]{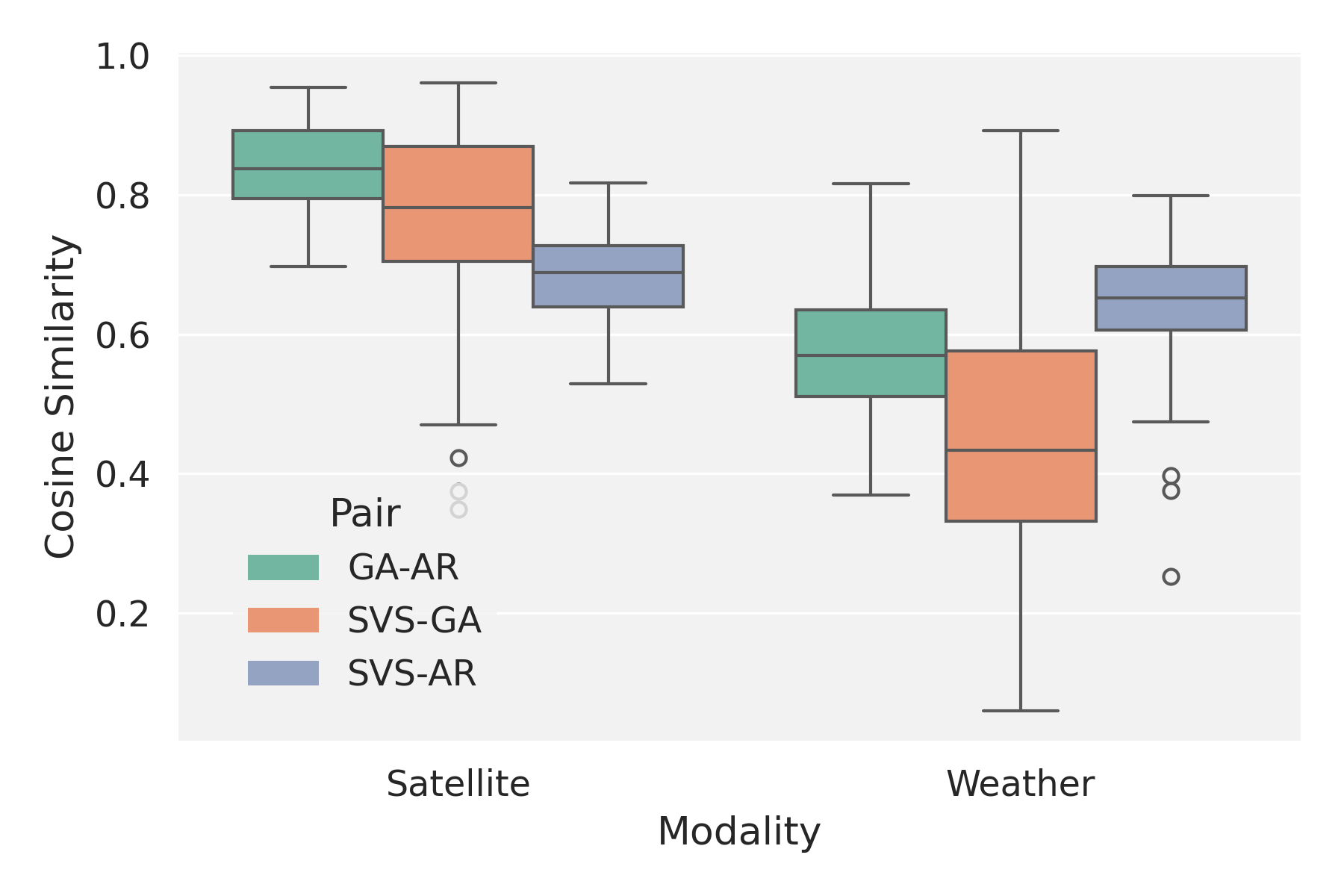}
                        \caption{Distribution of field-level cosine similarities between every pair of the compared attribution methods: \gls{ga}, \gls{ar} and \gls{svs}.
                        } 
                        \label{fig:cos_sim_boxplot}
                    \end{minipage}
                    \hspace{0.05\textwidth} 
                    \begin{minipage}{0.45\textwidth}
                        \centering  
                        \includegraphics[width=\linewidth]{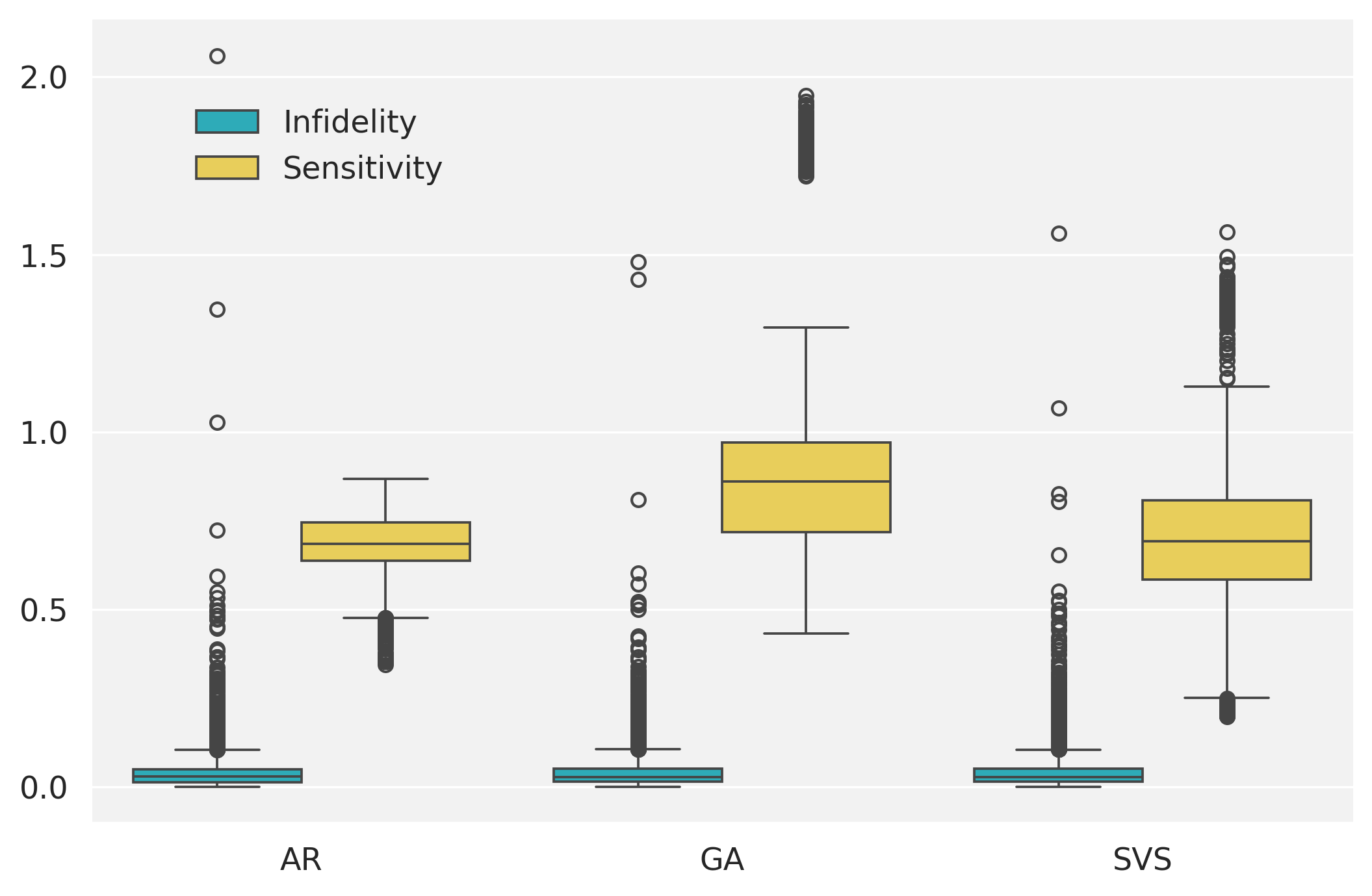}
                        \caption{Infidelity and Sensitivity scores of the temporal attributions estimated by \gls{ar}, \gls{ga}, and \gls{svs} methods.}
                        \label{fig:xai_metrics}
                    \end{minipage}
                \end{figure}

            \paragraph{Similarity Analysis}
                A visual assessment of the results in Figure \ref{fig:field_temp_attr} further reveals a degree of alignment among the attribution methods. 
                    For instance, there is a clear correspondence among the peaks identified by the three methods within the satellite attributions of Field-A, 
                    while a - less pronounced - alignment can be noticed in the weather attributions of Field-B, particularly between \gls{ar} and \gls{svs}.
                To quantitatively assess the similarity between the different methods, we calculate the cosine similarity between each pair using the field-level averaged attributions, and display the results in Figure \ref{fig:cos_sim_boxplot}. 
                    When comparing across modalities, we observe that the results of the satellite data consistently exhibits higher similarity scores than weather data. This suggests that the attribution methods align more closely when estimating temporal attributions for the satellite signal. A potential explanation lies in the lower entropy scores of weather data, which limits the temporal distribution of important information and increases the likelihood of alignment between methods.
                    When comparing across methods, \gls{ga} exhibits higher similarity to \gls{ar} than \gls{svs}, on average.  Interestingly, \gls{ar} and \gls{svs} provide the most similar weather attributions and the least similar satellite attributions. Yet, the similarity scores for this pair remain within a comparable range across both modalities, indicating a more robust and consistent alignment between \gls{ar} and \gls{svs}.
    
            \paragraph{Quantitative Evaluation}
               The sensitivity and infidelity metrics are used to evaluate and compare the robustness of the attributions generated by the three methods. Each metric assigns a single-valued score per pixel, with smaller values indicating greater robustness and stability. Figure \ref{fig:xai_metrics} presents these results across all corn fields, using 32 pixels per field. 
                    We observe that the infidelity scores are consistently low across all three methods, indicating a strong alignment between the magnitude of the attributions and the impact that input perturbations have on the model's predictions. It further suggests that all attribution methods effectively capture the relationship between input features and model outputs.
                    In contrast, the sensitivity scores are notably higher, particularly for the attributions generated by the \gls{ga} method. This result corresponds to the findings in subsection \ref{ssec:s2_scatterplot}, where \gls{ga} was observed to provide inconsistent and distant attributions for pixels within the same field. The stability of the \gls{ar} attributions is observed across other crops and regions, as we demonstrate in \ref{app:xai_metrics}.

\begin{figure*}[t]
                    \centering  
                    \adjustbox{center=\textwidth}{\includegraphics[width=1.35\textwidth]{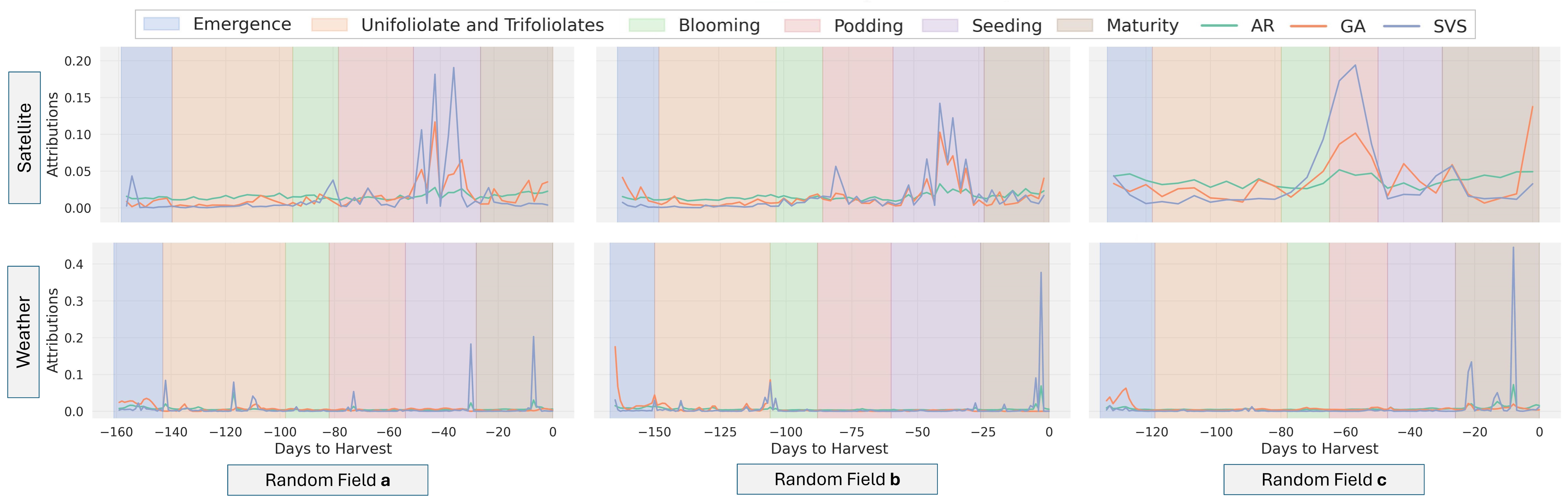}}
                    \caption{Field-level average attributions of the satellite and weather modalities, for three random soybean fields. The six growth stage periods are shown in the background of each plot. 
                    }
                    \label{fig:field_temp_attr_rdm_soybean_gs}
                \end{figure*}

            \paragraph{Agronomic interpretation}
                The attribution results can potentially offer valuable agronomic insights about the growth stages most influential for the model's predictions. Such analysis requires additional information about the start and end dates of the crop's growth stages at each field. While this metadata was not consistently measured throughout the growing season, we obtained approximate growth stage information for a subset of soybean fields\footnote{Phenology data was provided by \href{www.xarvio.com}{www.xarvio.com}, using their in-house developed and commercially deployed growth stage models.}.
                Figure \ref{fig:field_temp_attr_rdm_soybean_gs} illustrates three examples of soybean fields, where the temporal attributions are presented alongside the corresponding growth stages. 
                To address \ref{rq4}, we study the alignment between the identified critical growth stages and established agronomic knowledge, and provide a detailed analysis in \ref{app:temp_attr_soybean_gs}. Further comments on the results are discussed in the Discussion section \ref{sec:discussion}. 
                
        \subsection{Weather Events}\label{ssec:weather_events}
            To investigate the possible impact of special weather events on their attributions, we train a decision tree model to predict the attribution of each time step based on its weather properties: minimum, average, and maximum daily temperatures, as well as total precipitation. 
            We additionally include the number of days before harvest among predictive features, allowing the model to contextualize each weather event within the growth cycle of the crop.
            Specifically, we randomly sample 200 pixels from each field, merge the associated weather time series together, shuffle the instances to break the sequences, and then partition the datasets into 80\% for training and 20\% for testing.

            We train a separate decision tree for each set of corn fields belonging to the same farm and the same year. 
            We experiment with decision tree depths of two and three, to ensure the learned models remain interpretable.

\begin{figure*}[t]
                \centering  
                \includegraphics[width=0.7\linewidth]{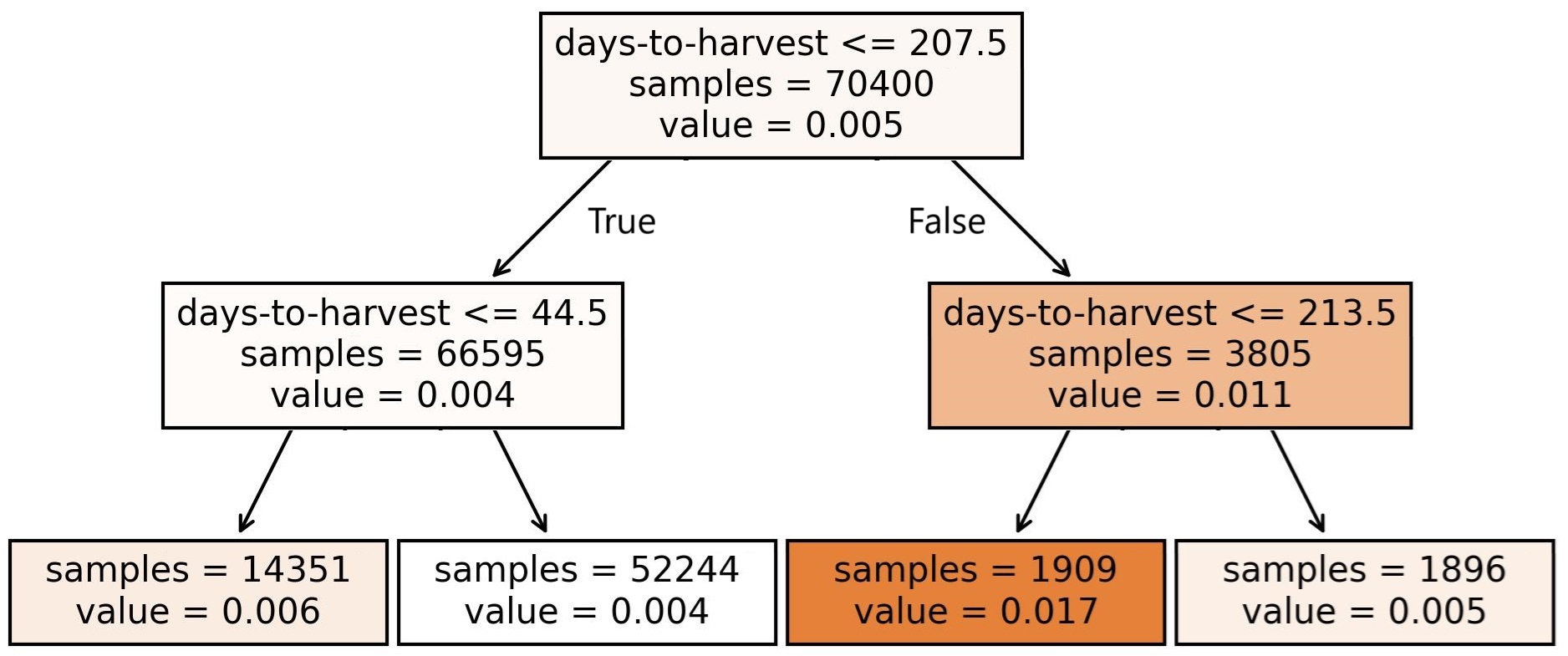}
                \caption{Decision Tree with two levels, predicting the \gls{ar} temporal attributions of the weather Transformer encoder. The color of each box is used as a scale for the predicted attribution values.
                }
                \label{fig:weather_DT2}
            \end{figure*}

            Figure \ref{fig:weather_DT2} presents the results of a two-level decision tree model trained on data from a farm with two fields from the 2023 season, using \gls{ar} scores as the target attributions. The \gls{r2} scores of the decision tree for this farm were particularly high and thus reliable for interpretation, reaching 0.75 in both the training and test sets. The figure represents the splitting of the training set, composed of 70,400 samples.
            We observe that the number of days before harvesting is the only variable used to split the tree.
            The darkest leaf in the tree, representing 2.7\% of the training set (1909 samples), shows a notably high attribution score of 0.017. These high-importance events occur between 213 and 207 days before harvesting. The remaining instances have attributions between 0.004 and 0.006, covering 97.3\% of the training samples. These results reveal where the highest mass of the attributions is located within the growth cycle for a specific farm.

            In \ref{app:weather}, we extend the analysis using a three-level decision tree trained on a different farm, where weather variables are utilized for splitting criteria.  
            Such analysis is generally useful in identifying weather events that significantly influence the decisions made by the Transformer model, highlighting the critical role that specific temperature conditions might play during particular days of the crop growth period.

    \subsection{Modality Importance}\label{ssec:mod_imp}

\begin{figure*}[t]
                \centering
                \includegraphics[width=0.8\linewidth]{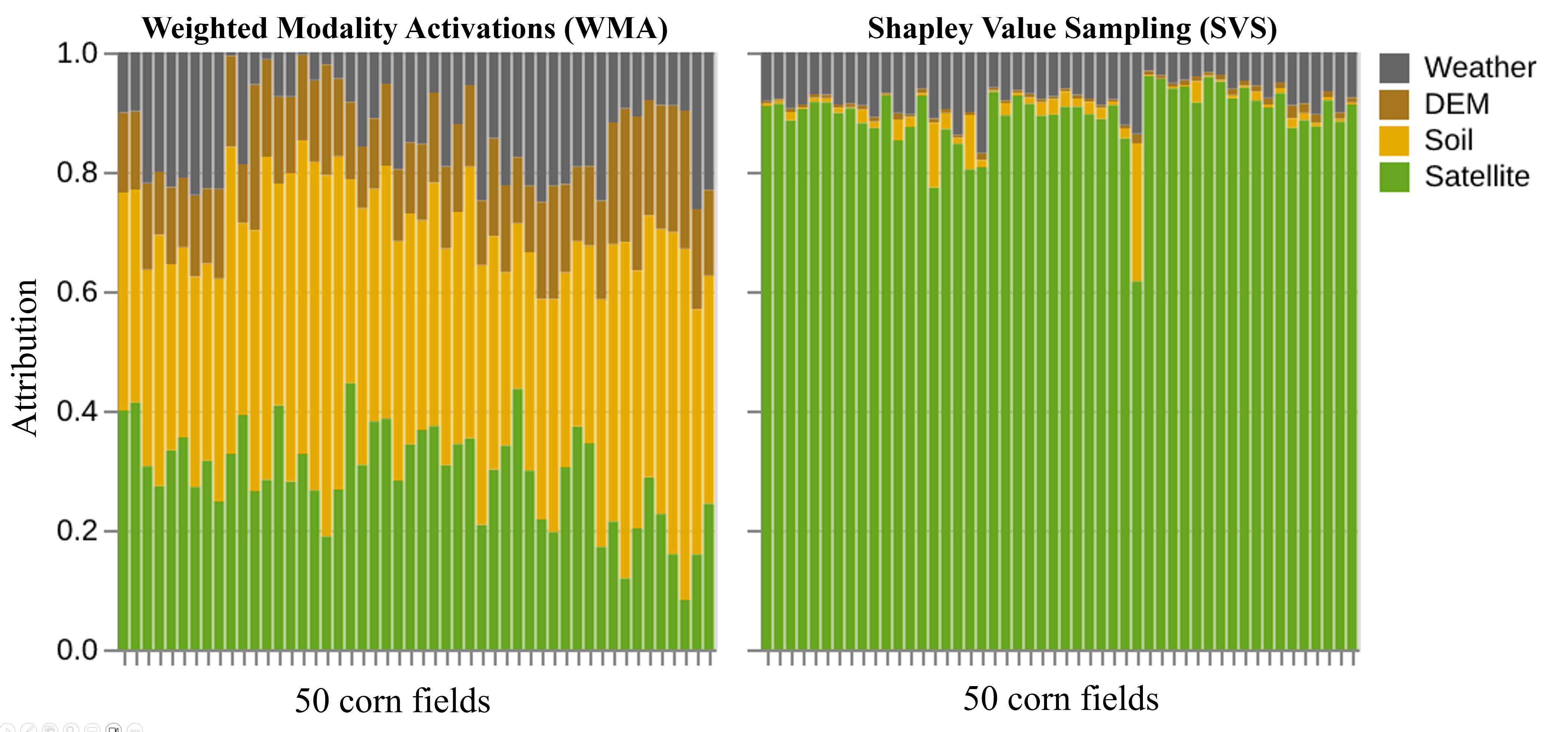}
                \caption{
                    Comparing the modality scores for the same random set of 50 corn fields. 
                 }
                \label{fig:modality_imp_corn}
            \end{figure*}

            After evaluating temporal attributions for satellite and weather modalities in Section \ref{ssec:temp_attr} and analyzing their role in identifying important weather events in Section \ref{ssec:weather_events}, we now address \ref{rq5} and assess the attributions of entire modalities using \gls{svs} and \gls{wma} methods described in Section \ref{ssec:xai_mod}.
            We compute both scores on a random selection of 32 pixels per field, using the same pixels sampled in Section \ref{ssec:temp_attr}. The scores are then aggregated per field by averaging the modality scores across the 32 pixels. In Figure~\ref{fig:modality_imp_corn}, we compare the results of both methods and present the modality scores for 50 corn fields.

            The \gls{wma} scores indicate that soil features have the highest impact on the prediction, accounting for an average of 41.3\% across all fields, followed by satellite data at 29.4\%. Terrain elevation features and weather data have the smallest share of contributions, with an average importance of 15.1\% and 14.2\%, respectively.
            In contrast, Shapley values indicate a different distribution of relative importance, with satellite data contributing the predominant share at 89.5\% on average, followed by weather at 7.9\%. 
            Soil and \gls{dem} features have only a minimal impact, contributing less than 2\% and 1\%, respectively.
            
            We attach in \ref{app:mod_imp} the results of the same comparison for other crops and regions, in which the satellite modality remains predominant according to the \gls{svs} results, while it contributes much less according to the \gls{wma} scores.
            This difference can be particularly attributed to the computational process: \gls{wma} relies only on the regression head to infer modality scores, while the \gls{svs} method uses the entire model.

\section{Discussion}\label{sec:discussion}
    In this section, we respond to the research questions we addressed to summarize the main takeaways of our work. \\[0.1cm]

    \textbf{\textit{(\ref{rq1}) Why the Transformer-based model was chosen?}}
        To process multimodal data, we designed networks with an intermediate fusion mechanism, enabling the training of modality-specific encoders to address the unique characteristics of each modality effectively. 
        We evaluated convolutional, recurrent, and Transformer-based networks, comparing their accuracy and inference times. The results demonstrated that the Transformer-based model offers a good balance between high performance and inference speed, in addition to its potential to provide intrinsic interpretability of its predictions.\\[0.1cm]

    \textbf{\textit{(\ref{rq2}) What did the analysis of the intermediate representations reveal?}}
        The linear probing experiments revealed that satellite representations exhibit a significantly stronger linear correlation with the predictions compared to representations from other modalities. This linearity improves progressively across deeper layers of the model.
        Further analysis of the learned attention weights within the satellite encoder showed that the first three Transformer layers generate similar weights for pixels within the same fields, while the fourth layer introduces greater diversity at the field level. Importantly, these variations in attention weights were found to be uncorrelated with differences in predicted yield. 
        Examining the temporal distribution of attention weights uncovered distinct patterns of key time periods for the satellite and weather encoders. Entropy analysis highlighted that important weather information is concentrated within a few critical time steps, while the satellite information relevant for predictions is distributed more evenly across the entire growth cycle.\\[0.1cm]

    \textbf{\textit{(\ref{rq3}) Which method for estimating temporal attributions is most reliable?}}
        We compared two model-specific methods, namely \gls{ar} and \gls{ga}, with a model-agnostic technique, \gls{svs}. Due to the impracticality, or even unfeasibility, of acquiring ground truth labels for the feature importance scores at each pixel, we relied on comparative analyses and quantitative evaluation metrics to assess the attribution methods.
        The infidelity and sensitivity scores highlighted that \gls{ar} delivers more consistent and robust attributions compared to the \gls{ga} method. While \gls{svs} demonstrated performance close to \gls{ar}, we would favor the latter for two primary reasons:
            First, the intrinsic nature of \gls{ar} enhances its faithfulness to the model, ensuring the attributions are more aligned with the model's internal mechanisms.
            Second, \gls{ar} has significantly faster computation times compared to the lengthy calculations and perturbations required by the Shapley-based method.
        These factors collectively make \gls{ar} a preferable choice for interpreting model predictions in this context.\\[0.1cm]

    \textbf{\textit{(\ref{rq4}) Can the temporal attributions provide agronomically relevant insights?}}
        We acquired crop phenology information for certain soybean fields in Argentina and demonstrated how to interpret temporal attributions in the context of agronomic knowledge. As detailed in \ref{app:temp_attr_soybean_gs}, the availability of starting and ending dates for each growth stage in each field allowed us to validate the model’s reasoning against established expert insights.
        Specifically, we examined three soybean fields and showed how certain growth stages known to be critical for the soybean yield were also important for the model, and explained that certain visual indicator of the yield at certain stages might also have been used by the model for its predictions. In contrast, some critical growth stages appeared to have little influence on the model's decisions. This raises questions about potential gaps in the model’s reasoning, calling for further experiments across multiple fields to validate these observations and explore their implications for model performance.\\[0.1cm]
        
    \textbf{\textit{What is the utility of weather events' analysis?}}
        Temporal attributions of the weather data highlight the significance of specific time periods for the model’s predictions. The analysis of weather events conducted in subsection \ref{ssec:weather_events} explores whether particular weather events have a substantial influence on the model's decisions. 
        From the two-level decision tree examined in this subsection, the findings suggest that attribution scores are more dependent on temporal factors than on climate conditions. In contrast, the three-level tree-based model studied in \ref{app:weather} incorporate temperature levels among its key splitting criteria, revealing the correlation between highly important time steps and their weather properties.\\[0.1cm]
    
    \textbf{\textit{(\ref{rq5}) Which modality attribution method is most reliable?}}
        Shapley values stand out for their ability to capture feature interactions by employing principles from game theory, considering multiple feature subsets and their contributions to the model before inferring feature attributions. Notably, \gls{svs} results exhibit a stronger alignment with the linear separability observed in the linear probing analysis, which indicated that satellite representations were most correlated with the final prediction.
        In contrast, the strength of \gls{wma} scores lies in their inherent connection to the model's architecture, which makes their importance estimations more faithful to the model's behavior \citep{rudin2019stop}.
        Evaluating the correctness of these methods remains challenging, as the modality impact scores do not necessarily reflect the agronomic significance of each modality, where established field knowledge could have been leveraged as a reference. Instead, these scores indicate how the model uses each modality, which depends on its learning scheme and the data patterns it captured during training.

\section{Conclusion}\label{sec:conclusion}
    This work highlights the potential of leveraging intrinsic interpretability within transformer-based models to enhance understanding in multimodal learning frameworks. 
    We examined the learned representations for each modality, inferred temporal attributions using both model-specific and model-agnostic approaches, and proposed \gls{wma}, an intrinsic method to derive modality importance scores.
    Our experiments revealed the \gls{ar} as a reliable intrinsic method for estimating temporal attributions, outperforming both \gls{svs} and \gls{ga} methods. In contrast, the modality contributions evaluated by \gls{svs} indicated that satellite data has a predominantly high influence on the predictions, whereas \gls{wma} suggests a more evenly distributed contribution across the four input modalities.

    A notable limitation of this study arises from the variability in seeding and harvesting dates across different fields. This variability complicates the comparison of temporal attribution results at the dataset level, as the sequence lengths of the temporal modalities varies between fields. Adding to this challenge is the missing phenology information of various growth stages for most fields. Furthermore, the modality attribution analysis did not yield relevant insights due to the conflicting results obtained between \gls{wma} and \gls{svs} estimations.

    Future work should prioritize a detailed analysis of the modality attribution methods to explain the conflicting results, and a quantitative evaluation to determine the most reliable approach for assessing the relative importance of different modalities in yield prediction tasks. Additionally, obtaining detailed growth stage data is essential for extending agronomical analyses across multiple fields and deriving generalizable insights.   
    Ultimately, resolving challenges related to the interpretability can facilitate building on the explainability findings to enforce certain rules or constraints during the learning phase, potentially optimizing the model performance.
    We hope our findings trigger further studies to identify robust interpretation techniques tailored for multimodal learning, facilitating the deployment of trustworthy models in critical, data-rich domains such as environmental and agricultural monitoring.

\section{Declaration of competing interest}
    The authors declare that they have no known competing financial interests or personal relationships that could have appeared to influence the work reported in this paper.

\section{Acknowledgments}
    We acknowledge with gratitude the thoughtful feedback from Professor Ribana Roscher on an earlier draft of this work.
    
    H.Najjar acknowledges support through a scholarship from the University of Kaiserslautern-Landau.
    
    The research results presented are part of a large collaborative project on agricultural yield predictions, which was partly funded through the ESA InCubed Programme (\url{https://incubed.esa.int/}) as part of the project AI4EO Solution Factory (\url{https://www.ai4eo-solution-factory.de/}).

\section{Declaration of generative AI and AI-assisted technologies in the writing process}
    During the preparation of this work, the authors used LanguageTool and ChatGPT in order to improve readability and correct potential grammar mistakes and typos. After using these tools, the authors reviewed and edited the content as needed and take full responsibility for the content of the publication.

\printglossary[type=\acronymtype]

\newpage
\appendix

\section{Model Selection}\label{app:model_ft}
    
    \subsection{Hyperparameters finetuning}\label{app:model_ft1}

        To optimize model performance, we experimented with different hyperparameter configurations for each encoder architecture on the corn dataset.
        Specifically, for Transformer-based encoders, we experimented with different attention heads (1, 2, or 4), layers (2, 4, or 6), and hidden sizes (32 or 64 neurons in each linear layer within each layer and head). For both \gls{1d-cnn} and \gls{lstm} encoders, we tested output channels (and hidden sizes for the \gls{lstm}) of 32, 64, and 128 while varying the number of layers from 2 to 5.
        For the \gls{alstm} encoder, we kept the number of layers in the \gls{lstm} block fixed at 2 while evaluating different hidden sizes (32, 64, or 128). Each tested hidden size was applied consistently across different layers and encoders (i.e., the satellite and weather encoders).

        Prior to fine-tuning the model hyperparameters described above, we also explored different configurations for optimizers and learning rate scheduling on a single network. We tested three optimization methods: Stochastic Gradient Descent (SGD) \citep{sutskever2013importance}, Adam \citep{kingma2014adam}, and AdamW \citep{loshchilov2017decoupled}. We further tested three learning rate schedulers: fixed learning rate, reducing the learning rate on plateau, and linear warmup with cosine decay. The best \gls{r2} scores were obtained using the AdamW optimizer with the linear warmup and cosine decay scheduler. Further fine-tuning of the batch size, base learning rate, and weight decay for the best-performing optimizer yielded optimal performance with a batch size of 2048, a base learning rate of 0.001, and a weight decay factor of 0.02.

    \subsection{Selected models}\label{app:model_ft2}
        
        After evaluating model performance on the validation set, the models that achieved the highest \gls{r2} scores are as follows: the Transformer-based model with four attention heads, two layers, and a hidden size of 64 neurons; the \gls{1d-cnn} model with four layers, each containing 64 output channels; the \gls{lstm} model with three layers, each with a hidden size of 32; and the \gls{alstm} model with two \gls{lstm} layers, each containing 32 hidden neurons.
        For the explanation analysis, the selected Transformer model has a single attention head, a single layer, and a hidden size of 32.
        For complete details on model implementation, please refer to the accompanying code.

\section{Performance across years and farms}\label{app:lofo_loyo}  

    In our manuscript, we focus on yield simulation for historical records, splitting the data into training, validation, and testing sets. Yield forecasting scenarios might require alternative data splits, such as training on all available data while leaving one year out. Similarly, applying the model to new regions would necessitate excluding subregions from the training set. 

    While such scenarios typically require adjusted data collection and model fine-tuning, we nevertheless provide an outlook on the relative performance of our selected model for such applications. Specifically, we train and evaluate the same model architecture under \gls{loyo} and \gls{lofo} schemes, with training and validation splits adjusted accordingly. Results are reported in Tables \ref{tab:lofo} and \ref{tab:loyo}. Groups containing only a single field were excluded from the analysis.

    We observe a significant performance decline compared to the results in Table~\ref{tab:models_res}, which is expected since data collection and model fine-tuning were not optimized for the \gls{loyo} and \gls{lofo} scenarios.    
    Most year groups perform poorly, yielding negative \gls{r2} scores at both field and subfield levels. Notably, 2022 achieves the best performance despite its model being trained on the smallest dataset, as the year 2022 covers the largest number of fields among all years. However, this pattern is inconsistent across schemes: Table~\ref{tab:lofo} shows Farm 11 (with only two fields) outperforming Farm 01 (with 40 fields, the largest group in \gls{lofo}).
    While further analysis could clarify these disparities across years and regions, we recommend prioritizing the collection of well-balanced datasets before fine-tuning models for specific data-splitting schemes.

    \begin{table}[t]
    \caption{Evaluation of the validation sets of the \gls{loyo} experiments, sorted by the ascending order of the validation year.}
    \small
    \adjustbox{center=\textwidth}{
    \begin{tabular}{cccccccccccc}
    \hline
     & \multirow{2}{*}{\begin{tabular}[c]{@{}c@{}}Validation \\ group\end{tabular}} & \multirow{2}{*}{\begin{tabular}[c]{@{}c@{}}Number \\ of Fields\end{tabular}} &  & \multicolumn{3}{c}{Subfield-Level} &  & \multicolumn{3}{c}{Field-Level} &  \\ \cline{5-7} \cline{9-11}
     &  &  &  & R² & RMSE (t/ha) & MAE (t/ha) & \textbf{ } & R² & RMSE (t/ha) & MAE (t/ha) &  \\ \cline{2-3} \cline{5-7} \cline{9-11}
     & 2017 & 9 &  & 0.01 & 2.37 & 1.58 &  & -0.62 & 1.28 & 0.91 &  \\
     & 2018 & 4 &  & -0.37 & 2.46 & 1.88 &  & -1.71 & 1.95 & 1.5 &  \\
     & 2020 & 9 &  & -0.38 & 2.67 & 2.18 &  & -1.14 & 1.57 & 1 &  \\
     & 2021 & 21 &  & 0.09 & 3.03 & 2.3 &  & 0.29 & 1.68 & 1.23 &  \\
     & 2022 & 68 &  & 0.38 & 2.75 & 2.13 &  & 0.57 & 1.65 & 1.26 &  \\
     & 2023 & 35 &  & -0.37 & 3.28 & 2.65 &  & -1.42 & 2.81 & 2.29 &  \\ \hline
    \end{tabular}
    }
    \label{tab:loyo}
    \end{table}

    \begin{table}[t]
    \caption{Evaluation of the validation sets of the \gls{lofo} experiments, sorted by the descending order of the number of fields in each group.}
    \small
    \adjustbox{center=\textwidth}{
    \begin{tabular}{cccccccccccc}
    \hline
     & \multirow{2}{*}{\begin{tabular}[c]{@{}c@{}}Validation \\ group\end{tabular}} & \multirow{2}{*}{\begin{tabular}[c]{@{}c@{}}Number \\ of Fields\end{tabular}} &  & \multicolumn{3}{c}{Subfield-Level} &  & \multicolumn{3}{c}{Field-Level} &  \\ \cline{5-7} \cline{9-11}
     &  &  & \textbf{} & R² & RMSE (t/ha) & MAE (t/ha) & \textbf{} & R² & RMSE (t/ha) & MAE (t/ha) &  \\ \cline{2-3} \cline{5-7} \cline{9-11}
     & Farm 01 & 40 &  & 0.52 & 2.81 & 2.24 &  & 0.7 & 0.18 & 1.5 &  \\
     & Farm 02 & 26 &  & -0.07 & 2.45 & 1.86 &  & -0.12 & 0.12 & 0.66 &  \\
     & Farm 03 & 19 &  & -0.05 & 2.23 & 1.74 &  & -0.03 & 0.14 & 1.34 &  \\
     & Farm 04 & 18 &  & 0.12 & 2.26 & 1.72 &  & 0.37 & 0.12 & 0.64 &  \\
     & Farm 05 & 8 &  & 0.12 & 2.23 & 1.73 &  & -0.17 & 0.2 & 1.23 &  \\
     & Farm 06 & 6 &  & 0.18 & 2.4 & 2.01 &  & 0.16 & 0.26 & 1.42 &  \\
     & Farm 07 & 6 &  & 0.18 & 2.69 & 1.97 &  & -2.34 & 0.09 & 0.94 &  \\
     & Farm 08 & 4 &  & -0.06 & 2.03 & 1.5 &  & 0 & 0.11 & 0.9 &  \\
     & Farm 09 & 3 &  & 0.7 & 1.84 & 1.4 &  & -3.3 & 0.09 & 0.54 &  \\
     & Farm 10 & 3 &  & 0.21 & 1.65 & 1.37 &  & 0.13 & 0.12 & 1.07 &  \\
     & Farm 11 & 2 &  & 0.59 & 1.66 & 1.31 &  & 0.83 & 0.07 & 0.28 &  \\
     & Farm 12 & 2 &  & 0.34 & 1.55 & 1.28 &  & 0.53 & 0.09 & 0.44 & \\
    \hline
    \end{tabular}
    }
    \label{tab:lofo}
    \end{table}

\section{Attention weights distribution}\label{app:raw_att}     

\begin{figure*}[h]
            \centering
            \adjustbox{center=\textwidth}{\includegraphics[width=1.3\textwidth]{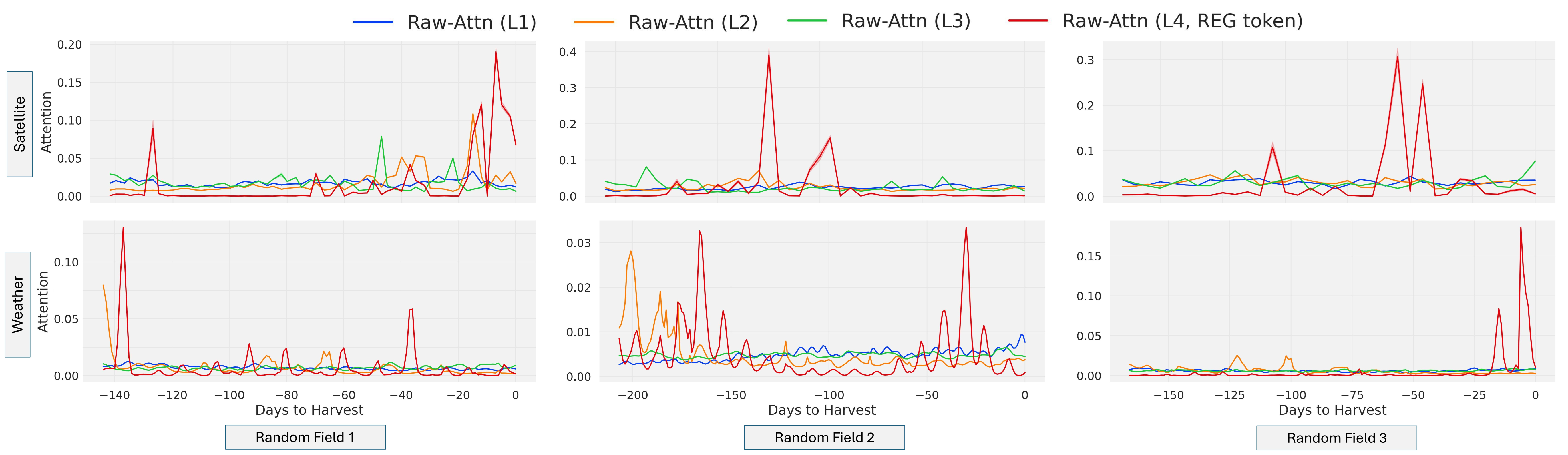}}
            \caption{
                Total attention weights attending at each time step for the first three attention layers, and the regression token weights in the final layer. 
                The results are averaged across 200 randomly selected pixels from three random fields, and are displayed for the satellite (a) and weather (b) Transformer encoders. 
                The light buffer regions represent the 95\% confidence interval around the average value.
                }
            \label{fig:att_layers_s2_weather_rdm}
        \end{figure*}

        \paragraph{Additional corn fields}\label{app:att_layers_rdm_fields}
        Figure \ref{fig:att_layers_s2_weather_rdm} displays the comparison of attention weights distribution across different layers of the Transformer encoder of satellite and weather modalities, for three random corn fields.
        
\section{Temporal Attributions}
    
    \subsection{Additional corn fields}\label{app:temp_attr_rdm_corn} 
\begin{figure*}[th]
            \centering  
            \adjustbox{center=\textwidth}{\includegraphics[width=1.3\textwidth]{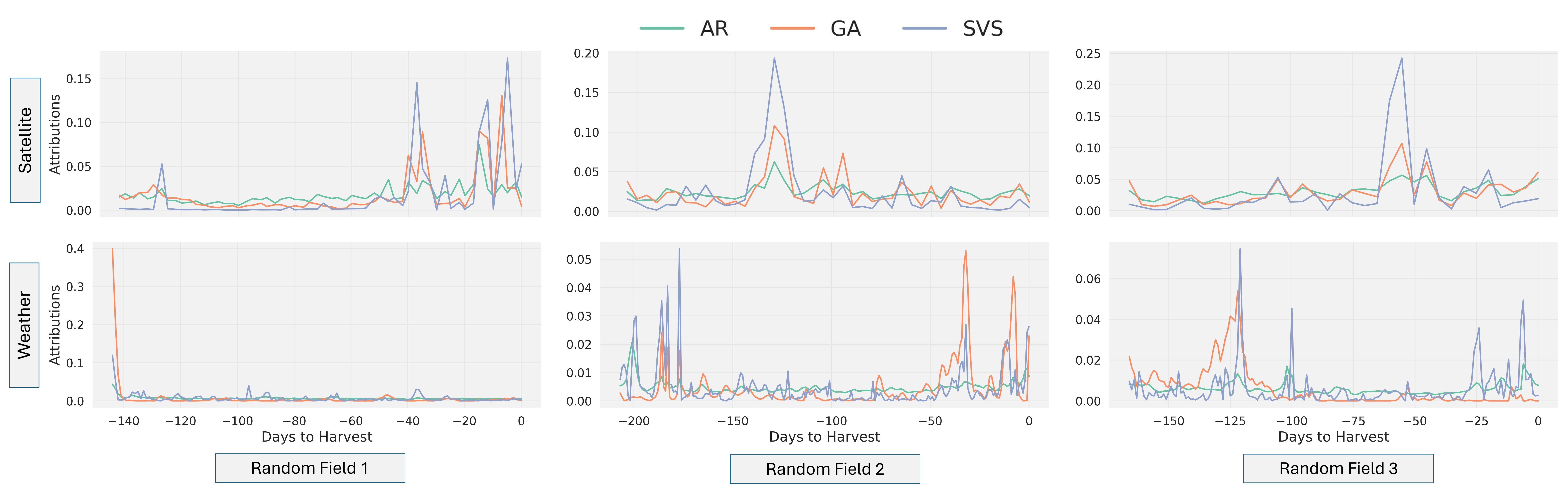}}
            \caption{Field-level average attributions of the satellite and weather modalities, for three random corn fields. 
            Due to the high computational cost associated with the \gls{svs} method, we limited the number of pixels sampled per field to 32 pixels. 
            }
            \label{fig:field_temp_attr_rdm}
        \end{figure*}

        In Figure \ref{fig:field_temp_attr_rdm}, we display the temporal attributions of the three evaluated methods for the same three random corn fields presented in \ref{app:raw_att}.
        
    \subsection{Quantitative Evaluation}\label{app:xai_metrics}
\begin{figure}[!t]
            \centering
            \adjustbox{center=\textwidth}{\includegraphics[width=1.2\linewidth]{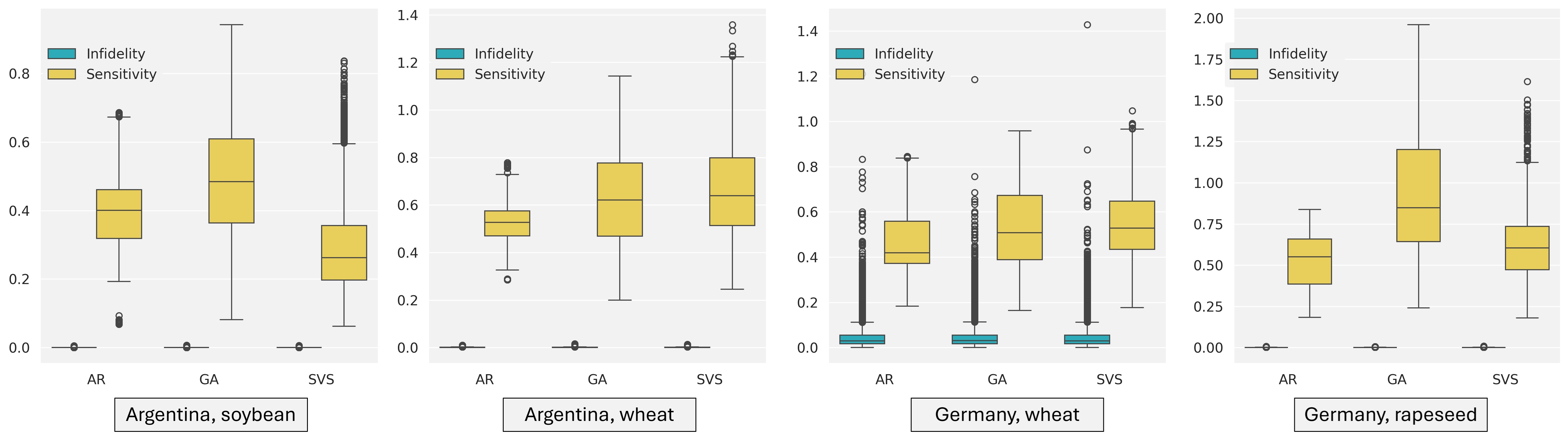}}
            \caption{Infidelity and Sensitivity scores of the temporal attributions estimated by AR, GA, and SVS methods, for soybean and wheat fields from Argentina, and wheat and rapeseed fields from Germany.}
            \label{fig:xai_metrics_other_data}
        \end{figure}

        We evaluate the methods for estimating temporal attributions using the infidelity and sensitivity scores on soybean, wheat and rapeseed fields in Argentina and Germany. The results displayed in Figure \ref{fig:xai_metrics_other_data} reveal that the infidelity scores are similarly low in all the datasets, while the sensitivity scores reflect varying ranks. \gls{ar} exhibits lowest sensitivity across all fields, except for soybean crops in Argentina, while \gls{ga} achieves comparable score as \gls{svs} in wheat fields, and a worse performance in rapeseed and soybean fields. Overall, a comparison of the two intrinsic methods reveals that \gls{ar} consistently provides more stable attributions compared to \gls{ga}.
        
    \subsection{Agronomic validation}\label{app:temp_attr_soybean_gs} 
        To illustrate how the temporal attributions can be interpreted in the light of agronomically meaningful periods, we retrieve some soybean fields for which we obtained approximated information about the start and end dates of different soybean growth stages.
        We overlap the phenological periods on the temporal attribution plots for three random soybean fields in Figure \ref{fig:field_temp_attr_rdm_soybean_gs}.
        The results display distinctive patterns across the modalities. We first analyze the satellite encoder:

        \begin{itemize}
            \item[$\circ$] Satellite data has a relatively low importance during the initial two stages of \textbf{emergence} and \textbf{leaf development} (i.e. unifoliolate and trifoliolates). 
            During this vegetation phase, the main stem nodes and their branching are developing, influencing the canopy structure and the final number of nodes \citep{kumudini2010soybean}. As a result, a poor canopy expansion might be identifiable in the satellite pixels, and thus used by the model to correlate with lower yield values. However, the results suggest that the model does not rely significantly on this cue.

            \item[$\circ$] We then observe that attribution values slightly increase at the \textbf{blooming} and \textbf{podding stages}, particularly in Field \textbf{c}. Defoliation of the plant during late blooming is, in fact, known to negatively affect yield \citep{mcwilliams1999soybean}. The total number of mature nodes and pods that develop during these two stages is also correlated with yield and may alter the field's landscape, thereby serving as an early indicator of potential yield.
            
            \item[$\circ$] In contrast, the model seems to rely more on the \textbf{seeding stage} in Fields \textbf{a} and \textbf{b}. In fact, leaf loss of 100\% has been shown to reduce yields by 80\% during early seeding \citep{mcwilliams1999soybean}. Another visual hint that can be captured by the model through the satellite data is the green seeds, which appear only during this stage.
            
            \item[$\circ$] The final \textbf{maturity stage} has a moderate level of importance across the three fields. This can be explained by the identifiability of the pods: 95\% of the pods on the main stem reach their mature pod color at this stage, which can potentially serve as a visual cue of the yield \citep{kumudini2010soybean}.

        \end{itemize}
        
        Weather data attributions exhibits different patterns, as it provides a different type of information to the model.
        An examination of 50 additional soybean fields revealed that the climate conditions during the early growing period and as the harvesting date approaches typically have the greatest influence on the model's predictions. More particularly:

        \begin{itemize}
            \item[$\circ$] During \textbf{emergence} stage, emergence speed is impacted by temperature and moisture conditions \citep{mcwilliams1999soybean}. In addition to soybeans being highly sensitive to salt, soil composition plays a critical role in nitrogen fixation and nutrient absorption, processes that occur during the early growing stages. Meanwhile, weather conditions are essential in regulating nutrient dynamics in the soil, affecting both nutrient availability and uptake by plants \citep{corwin2021climate,essa2002effect,tsekhmeistruk2021influence}. This explains the relatively high attribution values observed during this stage, particularly highlighted by the \gls{ga} method.

            \item[$\circ$] The \textbf{podding} is the most crucial period for seed yield, and any stress from \textbf{late podding} until \textbf{full seeding} causes more yield reduction than at any other time \citep{mcwilliams1999soybean}. Thus, if the weather conditions are unfavorable, it is usually recommended to compensate with appropriate irrigation strategies. However, the model does not appear to have caught important patterns connecting weather conditions during podding and seeding to the predicted yield. 

            \item[$\circ$] While stress during the \textbf{maturity stage} has almost no effect on yield, adequate weather conditions are required for the soybeans to dry and reach at least 15\% moisture to be ready for harvest \citep{mcwilliams1999soybean}. High drought might, however, cause significant losses. This weather-related potential influence might explain the high attribution values observed at this stage, especially highlighted by the \gls{svs} method.
        \end{itemize}

\begin{figure*}[t]
            \centering  
            \adjustbox{center=\textwidth}{\includegraphics[width=1.35\textwidth]{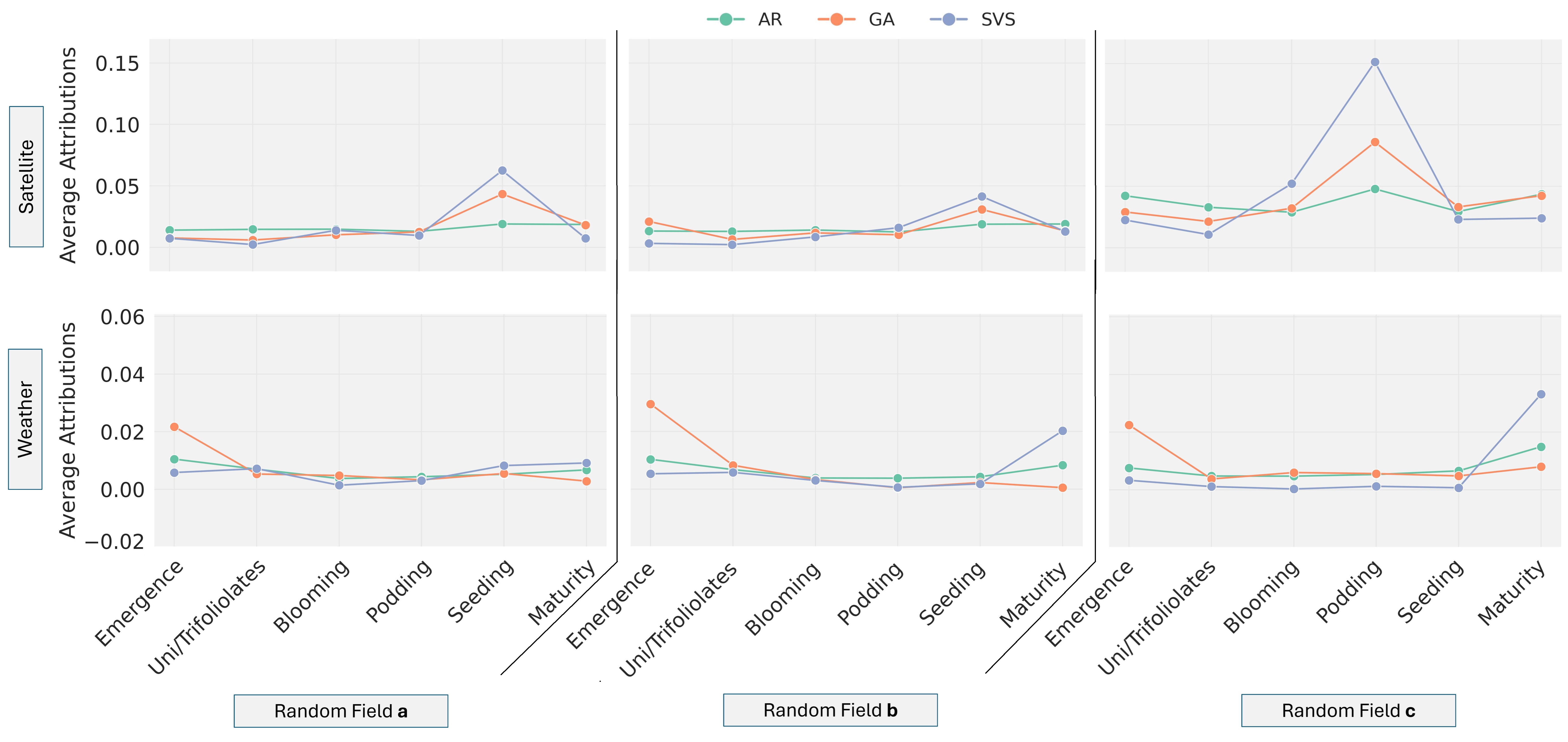}}
            \caption{Field-level average attributions of the satellite and weather modalities, further averaged per growth stage, for three random soybean fields.
            }
            \label{fig:field_avg_attr_rdm_soybean_gs}
        \end{figure*}

        Weather patterns remained consistent in most of the 50 soybean fields we examined. However, the temporal importance of satellite data showed more variance, predominantly fluctuating between the podding and seeding stages. 
        In Figure~\ref{fig:field_avg_attr_rdm_soybean_gs}, we aggregate the attributions per growth stage by averaging the values, and we observe similar patterns as the ones described above.
        Overall, this analysis highlights how some of the time steps that the model focuses on within each modality align with their corresponding agronomic significance, while other important patterns appear to be overlooked by the model. These observations, however, require further careful verification in collaboration with agronomy experts.

\section{Weather events}\label{app:weather}

\begin{figure*}[h]
        \centering  
        \includegraphics[width=\textwidth]{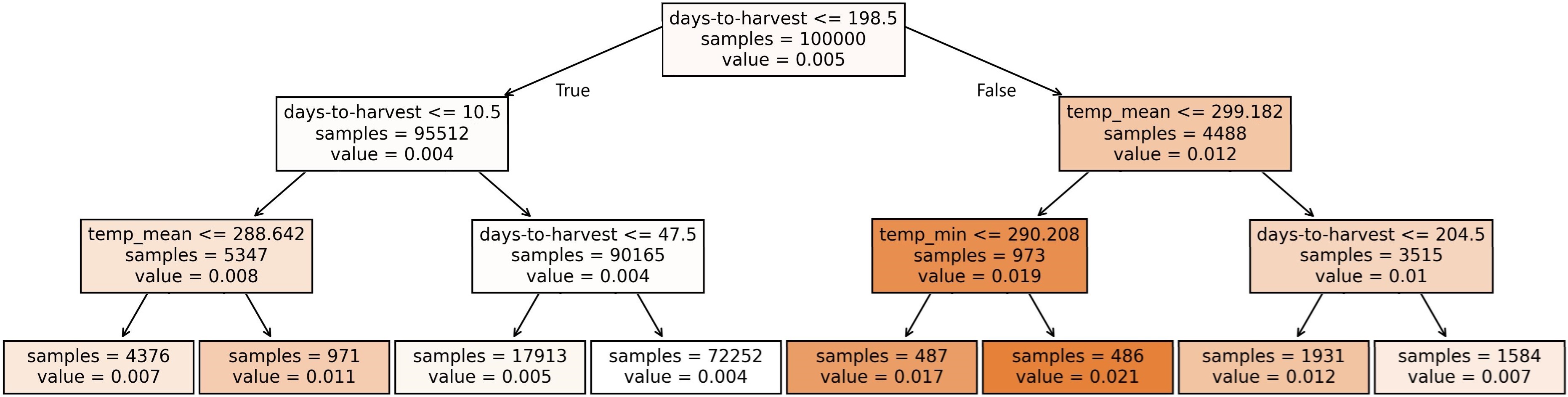}
        \caption{Decision Tree with three levels. The results shown are on the train set of 3 fields from the same farm, from 2023, predicting the \gls{ar} temporal attributions of the weather Transformer encoder.}
        \label{fig:weather_DT3}
    \end{figure*}

    We examine the correlation between particular weather events and their attributions by training decision trees.            
    We demonstrated in subsection \ref{ssec:weather_events} how interpreting a two levels-tree helped identify time periods with high importance to the model for a specific farm. Nevertheless, a slight increase in the tree depth can improve the tree performance across multiple farms while maintaining interpretability. Figure \ref{fig:weather_DT3} illustrates the weather events decision tree, of three levels, for a farm of three fields from the year 2023. For this farm, the tree model uses weather variables in addition to the number of days before harvesting of each instance. It achieves an accuracy of 79\% on both the training and test sets, on the task of predicting the \gls{ar} temporal attributions.
    
    We observe that the left branch of the tree covers a large portion of the training samples, greater than 95.5\%.  This branch includes 90\% of the instances with attribution scores ranging from 0.004 to 0.005, corresponding to days between 10 and 199 prior to harvesting. 
    To identify the samples most critical to the model's predictions, we focus on the nodes and leaves with the darkest shading. Approximately 10\% of the training samples occurring earlier than 198 days before harvesting, and associated with a mean temperature below 299.2 K, exhibited attribution scores between 0.017 and 0.021. Particularly, days where the minimum temperature exceeded 290 K attained the highest attribution score of 0.021.
    This finding indicates that such weather events are highly influential in the Transformer model, suggesting a critical role that specific temperature conditions play in the early days of the growing cycle.

\section{Modality importance}\label{app:mod_imp}

\begin{figure}[!t]
        \centering
        \includegraphics[width=0.75\linewidth]{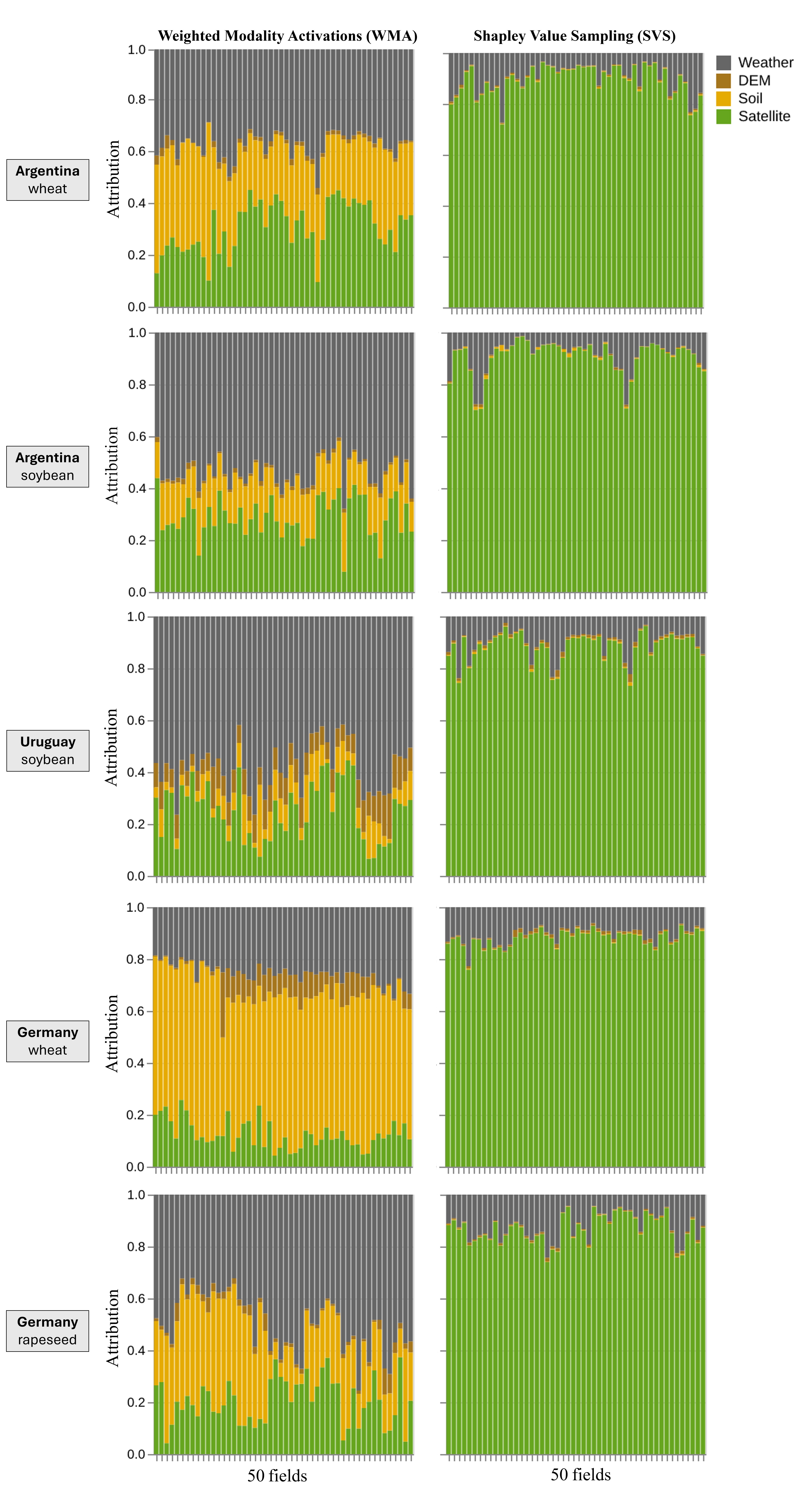}
        \caption{
            Comparing the modality importance using \gls{wma} and aggregated \gls{svs} scores for 50 fields of wheat and soybean in Argentina, soybean in Uruguay, and Wheat and rapeseed in Germany. 
         }
        \label{fig:modality_imp_others}
    \end{figure}

    \paragraph{Additional results} Figure~\ref{fig:modality_imp_others} compares the \gls{wma} and \gls{svs} scores for 50 random fields of soybean and wheat crops in Argentina, soybean in Uruguay, and wheat and rapeseed in Germany. 
    Consistent with the findings for corn fields shown in Figure \ref{fig:modality_imp_corn}, we observe that satellite data is the most influential modality according to Shapley-based scores, with terrain and soil having a marginal contribution.
    In contrast, the \gls{wma} scores suggest a reduced influence of satellite data, in favor of soil and weather modalities. Terrain elevation properties show minimal significance to the model across both interpretability techniques.

    When comparing crops across different regions, the modality scores for soybean fields are consistent between Argentina and Uruguay, with weather being the most influential modality, followed by satellite and then soil. 
    In contrast, models trained on wheat crops demonstrate a stronger reliance on soil and reduced usage of the satellite data in Germany compared to Argentina. 
    Overall, these regional differences likely reflect the impact of climate conditions, given that fields in Argentina and Uruguay are located in nearby regions and share similar climates, whereas German wheat crops grow in a different climatic environment than Argentinian wheat fields. 
    Additionally, climate impacts satellite data availability; for example, frequent cloudy weather in some regions renders several satellite images unusable for the model, further influencing the modality importance.
    Further analysis of how regional characteristics influence satellite data availability and quality could help explain the observed variations in modality attributions across datasets.

\newpage

\end{document}